\documentclass{article}

\usepackage{PRIMEarxiv}

\usepackage[utf8]{inputenc} 
\usepackage[T1]{fontenc}    
\usepackage{hyperref}       
\usepackage{url}            
\usepackage{booktabs}       
\usepackage{amsfonts}       
\usepackage{nicefrac}       
\usepackage{microtype}      
\usepackage{lipsum}
\usepackage{fancyhdr}       
\usepackage{graphicx}       
\graphicspath{{media/}}     

\usepackage{subcaption}
\usepackage{cite}
\usepackage{amsmath,amssymb,amsfonts}
\usepackage{algorithmic}
\usepackage{algorithm}
\usepackage{graphicx}
\usepackage{textcomp}
\usepackage{xcolor}

\usepackage[round]{natbib} 
\usepackage{multirow}

\title{Mixed-Precision Quantization for Deep Vision Models with Integer Quadratic Programming
\thanks{\hspace*{1mm}\textit{Accepted in Proceedings of the 62nd Annual Design Automation Conference. 2025.}} 
}

\author{
  Zihao Deng$^+$ \\
  \textit{The University of Texas at Austin} \\
  \texttt{zihaodeng@utexas.edu} \\
  \And
  Sayeh Sharify$^+$ \\
  \textit{d-Matrix Corp.} \\
  \texttt{sayehs@d-matrix.ai} \\
  \AND
  Xin Wang \\
  \textit{d-Matrix Corp.} \\
  \texttt{xwang@d-matrix.ai} \\
  \And
  Michael Orshansky \\
  \textit{The University of Texas at Austin} \\
  \texttt{orshansky@utexas.edu} \\
}

\begin{document}
\maketitle
\def\thefootnote{+} \footnotetext{\hspace*{1mm}These authors contributed equally to this work.}\def\thefootnote{\arabic{footnote}}

\begin{abstract}
Quantization is a widely used technique to compress neural networks. Assigning uniform bit-widths across all layers can result in significant accuracy degradation at low precision and inefficiency at high precision. Mixed-precision quantization (MPQ) addresses this by assigning varied bit-widths to layers, optimizing the accuracy-efficiency trade-off. Existing sensitivity-based methods for MPQ assume that quantization errors across layers are independent, which leads to suboptimal choices. We introduce CLADO, a practical sensitivity-based MPQ algorithm that captures cross-layer dependency of quantization error. CLADO approximates pairwise cross-layer errors using linear equations on a small data subset. Layerwise bit-widths are assigned by optimizing a new MPQ formulation based on cross-layer quantization errors using an Integer Quadratic Program. Experiments with CNN and vision transformer models on ImageNet demonstrate that CLADO achieves state-of-the-art mixed-precision quantization performance. Code repository available \href{https://github.com/JamesTuna/CLADO_MPQ}{here.\footnote{\hspace*{1mm}\url{https://github.com/JamesTuna/CLADO_MPQ}}}
\end{abstract}


\section{Introduction}

Reducing the storage and computational demands of deep neural networks (DNNs) is essential for practical applications. Model quantization effectively compresses DNNs~\citep{Courbariaux2015BinaryConnectTD, Nagel2020UpOD, Kim2021IBERTIB, Wei2022QDropRD, Chu_Li_Zhang_2024} and, when applied to pretrained models, is known as post-training quantization (PTQ). PTQ typically requires only a small dataset to calibrate activation and weight statistics, allowing efficient determination of optimal quantization ranges for each layer. This calibration process is both data-efficient and quick to deploy, with bit-widths specified for each layer.

Uniform precision quantization (UPQ), the simplest quantization approach, assigns the same bit-width to all layers but overlooks layer-specific tolerance to quantization. However, research has shown that certain layers are more robust to quantization than others~\citep{Dong2019HAWQHA, Cai2020ZeroQAN, Chauhan2023PostTM}; while quantizing some layers to low bit-width may yield minimal performance loss, even moderate quantization of other layers can lead to significant degradation. To take advantage of this phenomenon, mixed precision quantization (MPQ) optimizes bit-width assignments across layers such that lower prediction error is achieved at the same target compression rate as UPQ. Na\"ively formulated, the search space of MPQ grows \emph{exponentially} with the number of layers ($L$) to quantize. To address this, prior work presented two classes of solutions: \emph{search-based methods} and \emph{sensitivity-based methods}~\citep{Tang2022MixedPrecisionNN}. 

Search-based methods adopt search processes in which a large number of bit-width assignments are tested~\citep{Wu2018MixedPQ, Wang2019HAQHA, Lou2020AutoQAK, Guo2020SinglePO, Dong2023EMQET, azizi2024automated}. These search-based methods are usually expensive, hard to parallelize, and will take hundreds or even thousands of GPU hours because of their iterative search nature. Sensitivity-based methods, on the other hand, evaluate a closed-form metric that measures the tolerance of layers to quantization~\citep{Cai2020ZeroQAN, Dong2020HAWQV2HA, Ma2021OMPQOM, Chen2021TowardsMQ, Tang2022MixedPrecisionNN, Chauhan2023PostTM, yao2021hawq, kundu2022bmpq}.
Such a metric, \emph{viz.} \emph{sensitivity}, is a property of each quantized layer.
Various formulations have been proposed for the sensitivity metric in practice: \emph{e.g.} the Kullback-Leibler divergence between the quantized and the full-precision layer outputs~\citep{Cai2020ZeroQAN}, the largest eigenvalue of the Hessian~\citep{Dong2019HAWQHA}, the trace of the Hessian~\citep{Dong2020HAWQV2HA, yao2021hawq}, the Gauss-Newton matrix that approximates the Hessian~\citep{Chen2021TowardsMQ}, or the quantization scale factors~\citep{Tang2022MixedPrecisionNN}.
Despite the diversity in the formulation of sensitivity, all sensitivity-based MPQ methods minimize total sum of sensitivities across layers, constrained by a target compression ratio. Such an optimization problem is much more efficient to solve. These methods take several minutes to a few GPU-hours to complete the entire MPQ optimization process. Importantly, their runtime can be further reduced by parallel computation of sensitivities and these sensitivities can be reused when requirements change. \emph{Our work is also sensitivity-based}.

A key limitation of existing sensitivity-based MPQ algorithms is their assumption of independent layer-wise quantization effects, or dependence only within local blocks, as seen in BRECQ~\citep{Li2021BRECQPT}. This assumption allows for an objective function that is simple to formulate and optimize but overlooks the interactions between quantized layers—what we refer to as the \emph{cross-layer dependency of quantization error}. This oversight is especially impactful in computer vision models, where we demonstrate that an MPQ solution accounting for these dependencies outperforms methods assuming layer independence.

To overcome this limitation, we propose CLADO (\textbf{C}ross-\textbf{LA}yer-\textbf{D}ependency-aware \textbf{O}ptimization), an algorithm that measures the cross-layer dependency for all layers efficiently and optimizes a new cross-layer dependency aware objective via Integer Quadratic Programming (IQP). Following prior work, we start by applying the second-order Taylor expansion on network quantization loss.
Then, we decompose the loss into two components with each being a sum of terms.
Terms in the first component are specific to individual layers and they act independently across layers. They are well-addressed in prior work and we refer to them as the \emph{layer-specific sensitivities}.
Terms in the second component, which prior work ignores, represent interactions between layers and we refer to them as the \emph{cross-layer sensitivities}.
Na\"ively evaluating them requires calculation of inter-layer Hessian matrices, which is practically infeasible. We propose an efficient backpropagation-free method that avoids Hessian calculation and computes sensitivities by solving a set of linear equations that requires $O(L^2)$ times evaluations of the network on a small set of samples.
These sensitivities then allow us to reformulate the original MPQ problem as an IQP that contains $O(L)$ binary variables with linear constraints. We conduct experiments on multiple networks on the ImageNet dataset and demonstrate an improvement, in top-1 classification accuracy, of up to $27\%$ over uniform precision quantization, and up to $15\%$ over existing MPQ methods.
The contributions of our work are:

\begin{itemize}
    \item CLADO, the first sensitivity-based algorithm that captures the interactions of quantization error across \emph{all layers} and transforms the MPQ problem into an \emph{Integer Quadratic Program} that can be solved within seconds.

    \item CLADO is enabled by an efficient method to compute the cross-layer dependencies in $O(L^2)$ time.
    It is backpropagation-free and only requires forward evaluations of DNNs on a small amount of training data.

    \item 
    Experiments on computer vision models, including ResNets, MobileNets, and ViT, show that CLADO achieves state-of-the-art mixed-precision performance.
\end{itemize}

\section{Prior Work}
Existing mixed-precision quantization techniques can be divided into two classes: search-based and sensitivity based.

\textbf{Search-based methods:} These methods evaluate the performance of mixed-precision quantized networks and use it to guide a search process. HAQ~\citep{Wang2019HAQHA} and AutoQ~\citep{Lou2020AutoQAK} use Reinforcement Learning (RL), where all possible bit-width assignments of layers form the action space and the MPQ performance acts as the reward. Other search methods, such as MPQDNAS~\citep{Wu2018MixedPQ} and SPOS~\citep{Guo2020SinglePO}, apply Neural Architecture Search (NAS) to make MPQ a differentiable search process. 
The advantage of these methods is that they make decisions based on explicit evaluation of the quantized model's loss.
However, they are computationally demanding, usually taking hundreds or even thousands of GPU-hours to complete~\citep{Tang2022MixedPrecisionNN}.
Moreover, because each search is tied to a fixed set of constraints, when requirements change, new search processes need to be launched from scratch for search-based methods.

\textbf{Sensitivity-based methods:} These methods pre-compute \emph{layer sensitivity} as a proxy metric (or ``critic'') to assess the impact of quantization on model performance.
The sensitivities of layers in certain quantization precision are usually fast to measure, and once measured, they are reused in estimating the optimization objective as a function of different combinations of layer-wise precision assignments.
This type of methods, including ZeroQ~\citep{Cai2020ZeroQAN}, variants of HAWQ~\citep{Dong2019HAWQHA, Dong2020HAWQV2HA} and MPQCO~\citep{Chen2021TowardsMQ}, formulates the MPQ as a constrained combinatorial optimization problem.
The objective to be minimized is the sum of layer sensitivities that quantify the overall quantization impact, while the constraints ensure certain target compression requirements, such as memory and/or computational budgets. Other works, such as OMPQ~\citep{Ma2021OMPQOM}, BRECQ~\citep{Li2021BRECQPT}, and Bisection sensitivity-search~\citep{Chauhan2023PostTM}, also fall into the category of sensitivity-based methods and propose different approaches for measuring layer sensitivity. 
For layer sensitivities defined in some of the above mentioned work, EMQ~\citep{Dong2023EMQET} studied the consistency between them and the actual network performance.
We note that there is also prior work, BRECQ~\citep{Li2021BRECQPT}, that defines sensitivities on blocks of adjacent layers. BRECQ shares the same methodologies as the aforementioned approach in formulating the MPQ as an optimization problem with its objective being the summation of sensitivities and its constraints being the compression requirement.

To measure the layer sensitivities, a small subset of training data is needed (\emph{e.g.}, $1024$ training samples for ImageNet). We refer to it as the \emph{sensitivity set}. Since sensitivity-based methods optimize a proxy, instead of the real increase in network loss, they do not require iterative model evaluation and are efficient to solve. 
State-of-the-art sensitivity-based algorithms achieve performance comparable to that of search-based algorithms~\citep{Chen2021TowardsMQ}. \emph{Our proposed method CLADO is also sensitivity-based}.

\section{Cross-layer Dependency \& MPQ Optimality}
\label{sec:whyCross}
Empirically, cross-layer dependencies play a non-negligible role in the search of optimal MPQ solutions and ignoring them may lead to suboptimal solutions. Here we present, on ImageNet, two examples of MPQ suboptimality caused by ignoring the cross-layer dependencies. To make the suboptimality easy to understand, we focus on only a few layers and the goal is to \emph{choose two layers to quantize so that the quantization loss is minimized}.

\begin{figure}[!t]
    \centering
    \begin{subfigure}{.34\textwidth}
        \includegraphics[width=\textwidth]{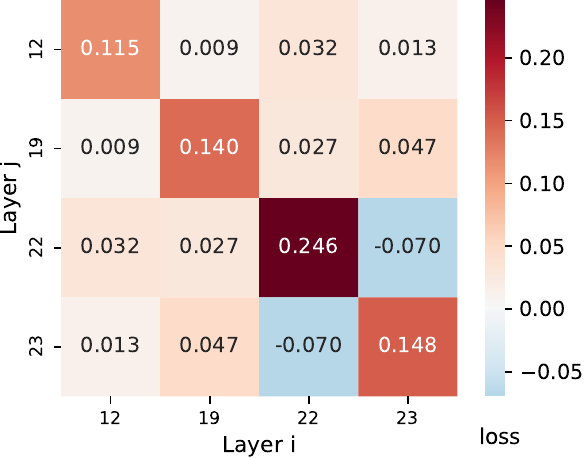}
        \caption{ResNet-34}
        \label{fig:sns-example-1}
    \end{subfigure}
    \begin{subfigure}{.32\textwidth}
        \includegraphics[width=\textwidth]{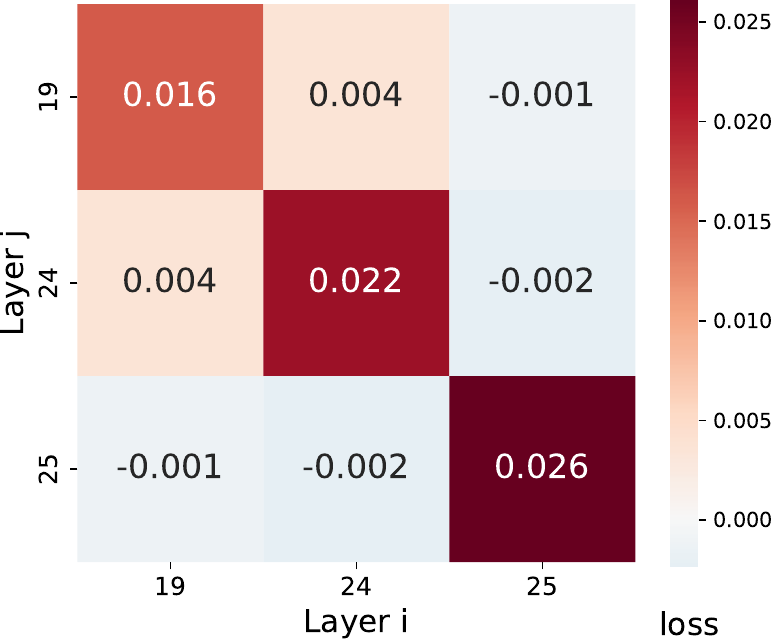}
        \caption{ResNet-50}
        \label{fig:sns-example-2}
    \end{subfigure}
    \caption{Sensitivity matrix of ResNet models. Entry $(i,i)$: increase in loss caused by quantizing a single layer $i$.
    Entry $(i,j)$: extra increase in loss when quantizing a pair of layers $(i,j)$, compared to individually quantizing them.}
\end{figure}


Figure~\ref{fig:sns-example-1} shows the sensitivity matrix of {ResNet-34} under 2-bit quantization for four specific layers: \texttt{layer2.2.conv2} (index 12), \texttt{layer3.1.conv2} (index 19), \texttt{layer3.3.conv1} (index 22), and \texttt{layer3.3.conv2} (index 23). Ignoring the cross-layer interactions (off-diagonal terms) leads to the selection of layers $(12,19)$ because it gives the smallest change in predicted loss $0.115+0.140=0.255$. However, the real induced loss change of quantizing layers $(12,19)$ is $0.115+0.140+2\times0.009=0.273$ and the optimal solution is to quantize layers $(22,23)$ because it has the smallest induced change in loss $0.246+0.148+2\times(-0.070)=0.254$.
Similarly, Figure \ref{fig:sns-example-2} shows the sensitivity matrix of ResNet-50 under 4-bit quantization for three layers: \texttt{layer2.2.conv3} (index 19), \texttt{layer3.0.conv2} (index 24), and \texttt{layer3.0.conv3} (index 25). Ignoring cross-layer interactions would lead to selecting layers $(19, 24)$, as this combination yields the smallest predicted loss change, calculated as $0.016+0.022=0.038$. However, the actual induced loss change is $0.038 + 2 \times 0.004 = 0.046$. The optimal solution instead is to quantize layers $(19, 25)$, yielding the smallest induced loss change of $0.016 + 0.026 + 2 \times (-0.001) = 0.040$.

BRECQ proposed a partial solution that addresses cross-layer dependencies for layers within residual bottleneck blocks~\citep{Li2021BRECQPT}. Measuring sensitivities of each block requires $|\mathbb{B}|^{L_{b}}$ measurements, where $L_{b}$ is the number of layers in the block. To scale BRECQ to account for cross-layer dependencies for all layers, $|\mathbb{B}|^{L}$ sensitivity measurements are needed. For many networks, it is practically infeasible. For example, considering two quantization bit-widths for ResNet-34, it takes around $3.4\times10^{10}$ measurements, which translates into $10^7\sim10^8$ GPU-hours assuming a typical sensitivity set of size 1024 (while our method takes only one GPU-hour\footnote{See the Experiment section for detailed results on runtime.}). Importantly, we show that ignoring inter-block dependencies is suboptimal in MPQ.\footnote{See the Ablation Study section for results and discussions.} Here we present the first sensitivity-based solution that efficiently addresses cross-layer dependencies over all layers.

\section{CLADO: Cross LAyer Dependency aware Optimization}
\label{sec:clado}

\subsection{Preliminaries}

We first introduce the formulation of the MPQ problem following Chen et al.~\citep{Chen2021TowardsMQ}.

\textbf{Notation:}
We assume an $I$-layer ({indices start from $0$}) neural network $f: \mathbf{\Theta} \times \mathbb{X} \rightarrow \mathbb{Y}$ and a training dataset of $N$ samples $\left(\mathbf{x}^{(n)}, \mathbf{y}^{(n)}\right) \in \mathbb{X} \times \mathbb{Y}$ with $n\in\{1, \ldots, N\}$. The model maps each sample $\mathbf{x}^{(n)}$ to a prediction $\hat{\mathbf{y}}^{(n)} = f\left( \theta, \mathbf{x}^{(n)} \right)$, using some parameters $\theta \in \mathbf{\Theta}$. Then the predictions are compared with the ground truth $\mathbf{y}^{(n)}$ and evaluated with a task-specific loss function $\ell: \mathbb{Y} \times \mathbb{Y} \rightarrow \mathbb{R}$. This leads to the objective function to minimize $\mathcal{L}: \mathbf{\Theta} \rightarrow \mathbb{R}$:
\begin{equation}
    \begin{aligned}
    \mathcal{L}(\theta) & =\frac{1}{N} \sum_{n=1}^N \ell\left(f\left(\theta, \mathbf{x}^{(n)}\right), \mathbf{y}^{(n)}\right) =\frac{1}{N} \sum_{n=1}^N \ell^{(n)}(\theta) 
    \end{aligned}
\end{equation}

We denote the weight tensor of the $i^{th}$ layer as $W^{(i)} \in \mathbb{R}^{c_o \times c_{in} \times k \times k}$ and its flattened version as $w^{(i)} \in \mathbb{R}^{c_o c_{in} k^2}$, where $k$ is the kernel size for a convolutional layer and $k=1$ for fully-connected layers, $c_{in}$ and $c_o$ are the number of input and output channels, respectively. The quantization function is denoted by $Q: \mathbb{R}^D \times \mathbb{Z}^{+} \rightarrow \Pi_b$, which takes a full-precision vector and the quantization bit-width as input and produces the quantized vector. Unless otherwise stated, we used per-tensor uniform symmetric quantization as it is the simplest scheme in practice due to its ease of implementation in hardware. As a result,  $\Pi_b=s \times\left\{-2^{b-1}, \ldots, 0, \ldots, 2^{b-1}-1\right\}$ for signed input and $s \times\left\{0, \ldots, 2^b-1\right\}$ for unsigned one, where $b$ is the quantization bit-width and $s$ is the quantization scale factor. The quantized $b$-bit weight is given by $Q(w, b)=$ $\operatorname{clip}\left(\lfloor w / s\rceil, -2^{b-1},2^{b-1}-1\right) \times s$ for full precision $w$.

\textbf{Discrete constrained optimization formulation:}
Let $w\triangleq\left\{w^{(i)}\right\}_{i=0}^{I-1}$ be the set of flattened weights of the network with $I$ layers. To find the optimal bit-width assignment with the goal of minimizing total model size, MPQ can be written as the following discrete constrained problem. $\mathbb{B}=\{b_1,b_2,...,b_{|\mathbb{B}|}\}$ denotes the set of bit-widths to be assigned in MPQ.
In this work, $C_{\text {target }}$ is the target model size of the network, and $|\cdot|$ denotes the length of vectors.
\begin{equation}
\begin{array}{c}
\min _{\left\{b^{(i)}\right\}} \frac{1}{N} \sum_{n=1}^N \ell\left(f\left(w+\Delta w, \mathbf{x}^{(\mathbf{n})}\right), \mathbf{y}^{(\mathbf{n})}\right) \\
\text { s.t. } \quad \Delta w^{(i)}=Q\left(w^{(i)}, b^{(i)}\right)-w^{(i)} \\
\sum_{i}\left|w^{(i)}\right| \cdot b^{(i)} \leq C_{\text {target }}\\
b^{(i)} \in \mathbb{B}, \,i \in \{0, \ldots, I-1\}\\
\end{array}
\label{eq:MPQForm}
\end{equation}

\textbf{Proxy of optimization objective:}
Let $g_w\triangleq \nabla \mathcal{L}(w)$ and $H_w\triangleq \nabla^2 \mathcal{L}(w)$ be the gradient and the Hessian, respectively.
For computational tractability, CLADO approximates the objective via the second-order Taylor expansion:
\begin{equation}
    \begin{aligned}
    \mathcal{L}(w+\Delta w) =\frac{1}{N} \sum_{n=1}^N \ell^{(n)}(w+\Delta w)\\
    \approx \mathcal{L}(w)+g_w^T \Delta w+\frac{1}{2} \Delta w^T H_w \Delta w 
    \end{aligned}
\label{eq:taylor1}
\end{equation}

For a well-trained model, it is reasonable to assume that training has converged to a local minimum, and thus, the gradient $g_w\approx 0$, resulting in:
\begin{equation}
    \begin{aligned}
    \mathcal{L}(w+\Delta w) \approx \mathcal{L}(w)+\frac{1}{2} \Delta w^T H_w \Delta w 
    \end{aligned}
\label{eq:taylor2}
\end{equation}

In essence, MPQ seeks to minimize $\mathcal{L}(w+ \Delta w)$ as a function of $\Delta w$, treating $\mathcal{L}(w)$ as a constant; thus, minimizing $\mathcal{L}(w+\Delta w)$ is equivalent to minimizing $\frac{1}{2}\Delta w^T H_w \Delta w$.
Here we use $\Delta w^T H_w \Delta w$ as a proxy of the objective in (\ref{eq:MPQForm}).

\subsection{Methods}
\label{sec:methods}
The core of CLADO is a way to capture in full the pairwise interactions of quantization error between all layers, which is either ignored, or only partially addressed in prior work.

\textbf{Layer-specific and cross-layer sensitivities:}
Let $H_{ij}$s be the partition of the Hessian at layer-level granularity.
Specifically, let
$H_{ii}\triangleq{\partial^2\mathcal{L}(w)}/{\partial {w^{(i)}} ^2}$ be the Hessian of layer $i$ and $H_{ij}\triangleq{\partial^2\mathcal{L}(w)}/{\partial w^{(i)} \partial w^{(j)}}$ be the cross-layer Hessian between layers $i$ and $j$.
Unless otherwise stated, we use $i,j$ to denote the indices of layers and they take values from $\{0,1,..,I-1\}$, where $I$ is the number of layers in the network. Then,
\begin{equation}
    \begin{aligned}
    \Delta w^T H_w \Delta w = \sum_{i}\Delta w^{(i)T}H_{ii} \Delta w^{(i)} + \sum_{i\neq j} \Delta w^{(i)T}H_{ij} \Delta w^{(j)}
    \label{eq:ObjDecomp}
    \end{aligned}
\end{equation}

Equation \ref{eq:ObjDecomp} decomposes the objective into two terms. The first term captures the contribution to loss increase from individual layers due to the intra-layer effects resulted from quantization.
We refer to them as \emph{layer-specific sensitivities}.
The second part
contains the contributions from effects of pairwise layer interaction as a consequence of two different layers being jointly quantized.
We refer to them as the \emph{cross-layer sensitivities}.
We use $\Omega_{i,j}(\cdot, \cdot)$ as a unified notation for both types of sensitivities.
\begin{equation}
    \begin{aligned}
    \Omega_{i,j}(\Delta w^{(i)},\Delta w^{(j)}) \triangleq &\Delta w^{(i)T}H_{ij} \Delta w^{(j)}
    \label{eq:sensitivityCLADO}
    \end{aligned}
\end{equation}

We rewrite (\ref{eq:ObjDecomp}) in terms of $\Omega$ as follows; minimizing (\ref{eq:ObjDecomp}) is equivalent to minimizing:
\begin{equation}
    \begin{aligned}
    \Omega=\sum_i \Omega_{i,i}(\Delta w^{(i)},\Delta w^{(i)})&+\sum_{i\neq j}\Omega_{i,j}(\Delta w^{(i)},\Delta w^{(j)})\\
    \end{aligned}
    \label{eq:newObjCLADO}
\end{equation}

Note that this re-definition obviates direct reference to the Hessian.
We will show below that $\Omega_{i,j}$ can be evaluated without computing the Hessian.
Unlike prior work~\citep{Dong2019HAWQHA,Cai2020ZeroQAN,Dong2020HAWQV2HA,Chen2021TowardsMQ} that ignores the cross-layer interaction terms, here we optimize (\ref{eq:newObjCLADO}) in its full form:
\begin{equation}
\begin{array}{c}
\min _{\left\{b^{(i)}\right\}} \Omega\\
\text { s.t. } \quad \Delta w^{(i)}=Q\left(w^{(i)}, b^{(i)}\right)-w^{(i)} \\
\sum_{i}\left|w^{(i)}\right| \cdot b^{(i)} \leq C_{\text {target }} \\
b^{(i)} \in \mathbb{B}, i \in\{0, \ldots, I-1\}
\end{array}
\label{eq:IQPForm2}
\end{equation}

\textbf{IQP formulation:}
Next we show that (\ref{eq:IQPForm2}) can be formulated as an Integer Quadratic Program (IQP) that solves for the bit-width assignment decisions. Since each layer $i$ has $|\mathbb{B}|$ candidate bit-width choices:$\{b_1,b_2,..,b_{|\mathbb{B}|}\}$, $\Delta w^{(i)}$ can take $|\mathbb{B}|$ values: $\Delta w^{(i)}\in\{\Delta w^{(i)}_{1},\Delta w^{(i)}_{2},..,\Delta w^{(i)}_{|\mathbb{B}|}\}$. Here, we use $\Delta w^{(i)}_m \triangleq Q(w^{(i)},b_m)-w^{(i)}$ with $m\in\{1,2,..,|\mathbb{B}|\}$ to denote the quantization error on $w^{(i)}$, when quantizing it to $b_m$ bits. For each layer $i$, we introduce a one-hot variable $\alpha^{(i)} \in \{0,1\}^{|\mathbb{B}|} \text{s.t.} \sum_{m=1}^{|\mathbb{B}|} \alpha_m^{(i)}=1$ to represent this layer's bit-width decision. The single entry of $1$ in $\alpha^{(i)}$, \emph{e.g.} $\alpha^{(i)}_m$, indicates the selection of a corresponding $\Delta w^{(i)}_m$ and the chosen bit-width $b_m$. For compact notation, let $\alpha\in\mathbb{R}^{|\mathbb{B}|I}$ be the concatenation of $\alpha^{(i)}$s, i.e., $\alpha=(\alpha^{(0)},\alpha^{},..\alpha^{(I-1)})$. Specifically,
\begin{equation}
\begin{aligned}
    &\Delta w^{(i)} = \alpha^{(i)}_1 \Delta w^{(i)}_1 + \alpha^{(i)}_2 \Delta w^{(i)}_2 + ... + \alpha^{(i)}_{|\mathbb{B}|} \Delta w^{(i)}_{|\mathbb{B}|}\\
    &\Delta w^{(i)}_m = Q(w^{(i)},b_m)-w^{(i)},\,m\in\{1,2,...,|\mathbb{B}|\}\\
    &\alpha^{(i)}_1,\alpha^{(i)}_2,...,\alpha^{(i)}_{|\mathbb{B}|} \in \{0,1\}, \alpha^{(i)}_1 + \alpha^{(i)}_2 + ... + \alpha^{(i)}_{|\mathbb{B}|} = 1\\
\end{aligned}
\label{eq:onehotvar}
\end{equation}

In addition, we gather all $\Omega_{i,j}$s into a matrix $\hat{G}\in\mathbb{R}^{|\mathbb{B}|I \times |\mathbb{B}|I}$, to which we refer to as the \emph{sensitivity matrix}. Specifically, for $m,n\in\{1,2,..,|\mathbb{B}|\}$:
\begin{equation}
\begin{aligned}
    \hat{G}_{|\mathbb{B}|i+m,|\mathbb{B}|j+n} \triangleq \Omega_{i,j}(\Delta w^{(i)}_m,\Delta w^{(j)}_n)\\
\end{aligned}
\label{eq:ltildeMatrix}
\end{equation}

Expanding the $\Delta w^{(i)}, \Delta w^{(j)}$ terms in (\ref{eq:ObjDecomp}) by (\ref{eq:onehotvar}) and bringing in the definition of $\hat{G}$ in (\ref{eq:ltildeMatrix}), we have $\Omega = \Delta w^T H_w \Delta w = \alpha^T \hat{G} \alpha$. Accordingly, the MPQ problem is equivalent to the following IQP:
\begin{equation}
\begin{array}{c}
\min _{\left\{\alpha^{(i)}\right\}} \alpha^T \hat{G} \alpha\\
\text { s.t. }
\alpha=(\alpha^{},\alpha^{(2)},..\alpha^{(I)})\\
\alpha^{(i)}_1,\alpha^{(i)}_2,...,\alpha^{(i)}_{|\mathbb{B}|} \in \{0,1\}, \alpha^{(i)}_1 + \alpha^{(i)}_2 + ... + \alpha^{(i)}_{|\mathbb{B}|} = 1\\
\sum_{i}\sum_{m}\alpha^{(i)}_m\left|w^{(i)}\right| \cdot b_m \leq C_{\text {target }} \\
i \in\{0, \ldots, I-1\} ,\,\, m\in\{1,\ldots,|\mathbb{B}|\}
\end{array}
\label{eq:IQPForm}
\end{equation}

\textbf{Backpropagation-free sensitivity measurement:}
We now present an efficient way to compute $\hat{G}$ through only forward evaluations of DNNs.
For any pair of layers $i,j$ and their corresponding bit-width choices $m,n$, computing $\Omega_{i}(\Delta w^{(i)}_{m})$, $\Omega_{j}(\Delta w^{(j)}_{n})$, and $\Omega_{i,j}(\Delta w^{(i)}_{m},\Delta w^{(j)}_{n})$ in a na\"ive manner requires computing the Hessians.
However, we point out that the Hessian computation can be avoided. We estimate $\Omega$ by directly measuring the change in loss due to network perturbations:
\begin{equation}
\begin{aligned}
&\Omega_{i,i}(\Delta w^{(i)}_{m},\Delta w^{(i)}_{m}) \approx 2 \left(\mathcal{L}(w+\Delta w^{(i)}_{m})-\mathcal{L}(w)\right)\\
&\Omega_{j,j}(\Delta w^{(j)}_{n},\Delta w^{(j)}_{n})  \approx 2 \left(\mathcal{L}(w+\Delta w^{(j)}_{n})-\mathcal{L}(w)\right)\\
&\Omega_{i,i}(\Delta w^{(i)}_{m},\Delta w^{(i)}_{m}) +  \Omega_{j,j}(\Delta w^{(j)}_{n},\Delta w^{(j)}_{n}) +2\Omega_{i,j}(\Delta w^{(i)}_{m},\Delta w^{(j)}_{n}) \\
&\approx 2 \left(\mathcal{L}(w+\Delta w^{(i)}_{m}+\Delta w^{(j)}_{n})-\mathcal{L}(w)\right) \\
\end{aligned}
\label{eq:FLDICT}
\end{equation}

Equations in (\ref{eq:FLDICT}) follow from equation (\ref{eq:taylor2}). The right sides of the three equations in (\ref{eq:FLDICT}) are computed by taking the difference in the loss of the quantized model from that of the full-precision model.
Subtracting the first two equations from the third equation:
\begin{equation}
\begin{aligned}
\Omega_{i,j}(\Delta w^{(i)}_{m},\Delta w^{(j)}_{n}) \approx \mathcal{L}(w+\Delta w^{(i)}_{m}+\Delta w^{(j)}_{n})+\mathcal{L}(w)
\\ - \mathcal{L}(w+\Delta w^{(i)}_{m}) - \mathcal{L}(w+\Delta w^{(j)}_{n}) 
\label{eq:FLDICT2}
\end{aligned}
\end{equation}

Importantly, (\ref{eq:FLDICT}) and (\ref{eq:FLDICT2})  compute all sensitivities by \emph{only forward evaluations} of DNNs on the small sensitivity set. In total, CLADO sensitivity computation requires $\frac{1}{2}|\mathbb{B}|I(|\mathbb{B}|I+1)$ measurements of the network on a small set of samples. After sensitivity computation, we apply a positive semi-definite (PSD) approximation to the matrix $\hat{G}$. Ideally, $G$ is PSD if computed on the entire training set after full convergence, but using a small sensitivity set can introduce measurement errors, making it indefinite. In practice, PSD approximation is crucial for meaningful MPQ solutions (see Section~\ref{sec:ablation} for details).

\begin{algorithm}[!t]
\caption{CLADO}
\label{alg:CLADO}
\begin{algorithmic}
\STATE {\bfseries Input:} $I$-layer network with bit-width options $\mathbb{B}=\{b_1,b_2,..,b_{|\mathbb{B}|}\}$ and model size target $C_{\text{target}}$.
\STATE {\bfseries Output:} The bit-width decision for each layer.
\item Evaluate the loss of full-precision model: $\mathcal{L}(w)$
\STATE Initialize $\hat{G}\in\mathbb{R}^{|\mathbb{B}|I\times |\mathbb{B}|I}$
\FOR {$i$ in $\{0,\ldots,I-1\}$}
    \FOR {$m$ in $\{1,\ldots,|\mathbb{B}|\}$}
        \STATE $\Delta w^{(i)}\gets Q(w^{(i)},b_m)-w^{(i)}$ \begin{small}\color{gray}{\hspace*{40mm}// Quantize layer $i$ to $b_m$ bits}\end{small}
        \STATE Evaluate loss of quantized network $\mathcal{L}(w+\Delta w^{(i)})$
        \STATE $S\gets2(\mathcal{L}(w+\Delta w^{(i)})-\mathcal{L}(w))$ \begin{small}\color{gray}{\hspace*{36.5mm}// Layer-specific sensitivity}\end{small}
        \STATE $\hat{G}_{|\mathbb{B}|i+m,|\mathbb{B}|i+m}\gets S$ \begin{small}\color{gray}{\hspace*{52mm}// Store layer-specific sensitivity}\end{small}
    \ENDFOR
\ENDFOR
\FOR {$i$ in $\{0,\ldots,I-2\}$}
        \FOR {$j$ in $\{i+1,\ldots,I-1\}$}
            \FOR {$(m,n)$ in $\{1,..|\mathbb{B}|\}\times\{1,..|\mathbb{B}|\}$}
                \STATE $\Delta w^{(i)}\gets Q(w^{(i)},b_m)-w^{(i)}$ \begin{small}\color{gray}{\hspace*{36mm}// Quantize layer $i$ to $b_m$ bits}\end{small}
                \STATE $\Delta w^{(j)}\gets Q(w^{(j)},b_n)-w^{(j)}$ \begin{small}\color{gray}{\hspace*{36mm}// Quantize layer $j$ to $b_n$ bits}\end{small}
               \STATE Measure loss of quantized model $\mathcal{L}(w+\Delta w^{(i)}+\Delta w^{(j)})$
                \STATE \begin{small}\color{gray}{// Compute cross-layer sensitivities using equation (\ref{eq:FLDICT2})}\end{small}
                \STATE $S\gets \mathcal{L}(w+\Delta w^{(i)}+\Delta w^{(j)})+\mathcal{L}(w)$ 
                \STATE $S\gets S-0.5\hat{G}_{|\mathbb{B}|i+m,|\mathbb{B}|i+m}$ 
                \STATE $S\gets S-0.5\hat{G}_{|\mathbb{B}|j+n,|\mathbb{B}|j+n}$ 
                \STATE $\hat{G}_{|\mathbb{B}|i+m,|\mathbb{B}|j+n}\gets S$
            \ENDFOR
        \ENDFOR
\ENDFOR
\STATE Compute the eigen-decomposition: $\hat{G} = \sum_{i=1}^{|\mathbb{B}|I}u_i e_i u_i^T$ and set PSD approximated $\hat{G}\gets \sum_{i=1,e_i>0}^{|\mathbb{B}|I}u_i e_i u_i^T$
\STATE Solve the IQP problem in (\ref{eq:IQPForm}) by plugging in $\hat{G}$,$\mathbb{B}$, and $C_{\text{tagret}}$ to get the optimal solution $\alpha^*$.
\STATE {\bfseries Return:} The bit-width decisions indicated by $\alpha^*$.

\end{algorithmic}
\end{algorithm}

\begin{table*}[!ht]
\footnotesize
\centering
\caption{MPQ results (PTQ): CNNs and ViT on ImageNet. Numbers denote the averaged Top-1 accuracy (percentage) for different algorithms. CLADO delivers the best MPQ results under most model sizes. (*: for ablation study, CLADO with cross-layer terms removed. $^+$: experiment using per-channel affine instead of per-tensor symmetric quantization.)}
\resizebox{0.98\textwidth}{!}{
\begin{tabular}{c|ccc|ccc|ccc|ccc|ccc}
    \toprule
    \multirow{3}{*}{Model} & \multicolumn{3}{|c}{ResNet-34} & \multicolumn{3}{|c}{ResNet-50}  & \multicolumn{3}{|c}{MobileNetV3-Large$^+$} & \multicolumn{3}{|c}{RegNet-3.2GF} & \multicolumn{3}{|c}{ViT-base$^+$}\\
    & \multicolumn{3}{|c}{INT8 size: 20.27MB;} & \multicolumn{3}{|c}{INT8 size: 22.36MB;} & \multicolumn{3}{|c}{INT8 size: 0.35MB;} & \multicolumn{3}{|c}{INT8 size: 13.57MB;} & \multicolumn{3}{|c}{INT4 size: 40.5MB;} \\
    & \multicolumn{3}{|c}{Acc: 73.10} & \multicolumn{3}{|c}{Acc: 75.87} & \multicolumn{3}{|c}{Acc: 72.20} & \multicolumn{3}{|c}{Acc: 77.88} & \multicolumn{3}{|c}{Acc: 79.47} \\
    
    \hline
    Size (MB)  & 10.13 & 12.16 & 14.19 & 10.00 & 13.42 & 17.89 & 0.21 & 0.25 & 0.29 & 6.78 & 9.50 & 12.89 & 29.84 &31.97 &34.11 \\
    \hline
    HAWQ & 56.86 &68.13 &72.00 & 53.70 &72.55 &75.36 &15.15 &44.88 &\textbf{71.82} &33.84 &73.08 &77.24 & 10.57 &52.44 &59.91 \\
    MPQCO & 53.39 &67.51 &71.53 & 45.90 &72.13 &74.64 &17.64 &67.62 &71.75 &29.34 &72.73 &77.51 & 60.96 &75.73 &77.15 \\
    CLADO* & 54.47 &70.37 &71.67 & 53.22 &72.51 &75.35 &43.80 &67.52 &71.75 &32.84 &73.48 &77.51 & 54.42 & 70.75 & 74.05 \\
    CLADO & \textbf{62.59} &\textbf{70.38} &\textbf{72.20} &\textbf{60.13} &\textbf{73.10} &\textbf{75.42} &\textbf{49.85} &\textbf{68.00} &71.81 &\textbf{41.40} &\textbf{73.98} &\textbf{77.53} & \textbf{66.60} & \textbf{76.15} & \textbf{77.88} \\
    \bottomrule
\end{tabular}
}
\label{tab:CLADO_PTQ_CNN}
\end{table*}

\section{Experimental Results: CNNs}
\label{sec:exp}
\subsection{Experimental Settings}
\textbf{Dataset and models:} We conduct experiments on the ImageNet dataset~\citep{Russakovsky2014ImageNetLS}.
To compare CLADO to existing work, we test two state-of-the-art algorithms: MPQCO~\citep{Chen2021TowardsMQ} and HAWQ-V3~\citep{yao2021hawq}. We evaluate with four convolutional models: ResNet-34/50~\citep{He2015DeepRL}, MobileNetV3~\citep{Howard2019SearchingFM}, and RegNet-3.2GF~\citep{Radosavovic2020DesigningND}. For all models excluding MobileNetV3, we quantize activations to 8-bit and consider three candidate bit-widths for weights: $\mathbb{B}=\{2,4,8\}$. (Note that, as shown in Algorithm~\ref{alg:CLADO}, CLADO can work with any number ($|\mathbb{B}|$) of bit-width candidates, for experiments, we use $|\mathbb{B}|=3$.) We consider more conservative quantization bit-width candidates $\mathbb{B}=\{4,6,8\}$ for MobileNetV3 because of its high parameter efficiency.

\begin{figure}[!t]
    \centering
    \begin{subfigure}{.32\textwidth}
        \includegraphics[width=.99\textwidth]{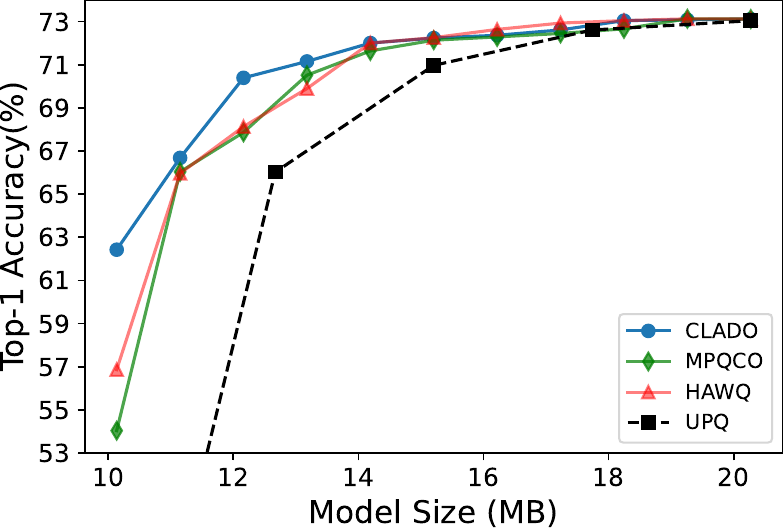}
        \caption{ResNet-34}
    \end{subfigure}
    \begin{subfigure}{.32\textwidth}
        \includegraphics[width=.99\textwidth]{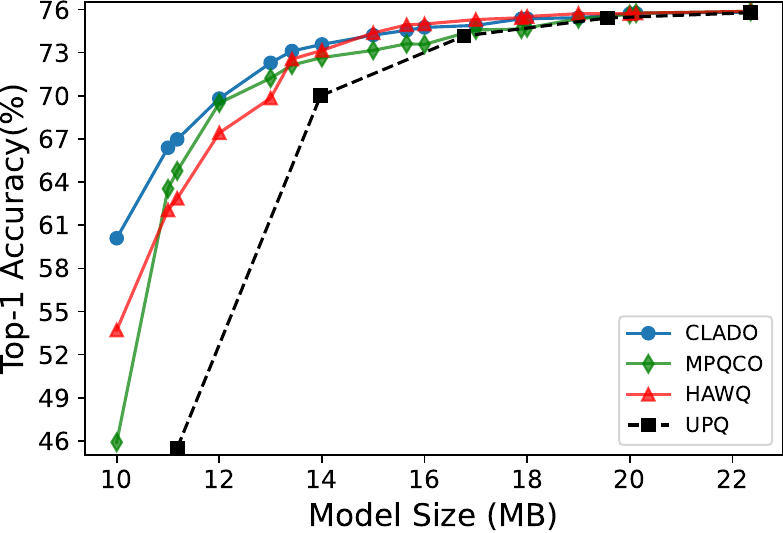}
        \caption{ResNet-50}
    \end{subfigure} 
    \begin{subfigure}{.33\textwidth}
        \includegraphics[width=.99\textwidth]{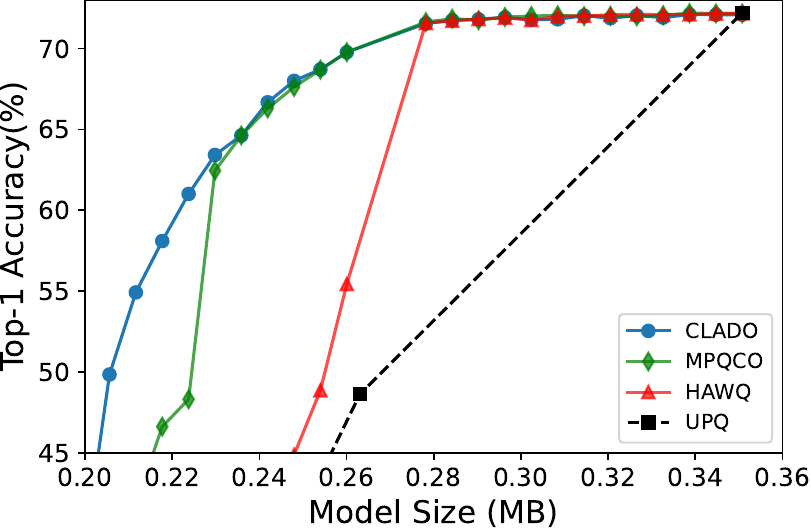}
        \caption{MobileNetV3}
    \end{subfigure} \\
    \begin{subfigure}{.32\textwidth}
        \includegraphics[width=.99\textwidth]{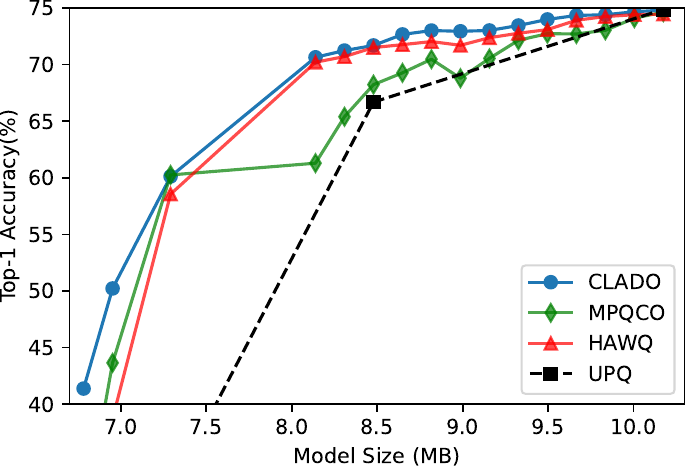}
        \caption{RegNet-3.2GF}
    \end{subfigure} 
    \begin{subfigure}{.33\textwidth}
        \includegraphics[width=.99\textwidth]{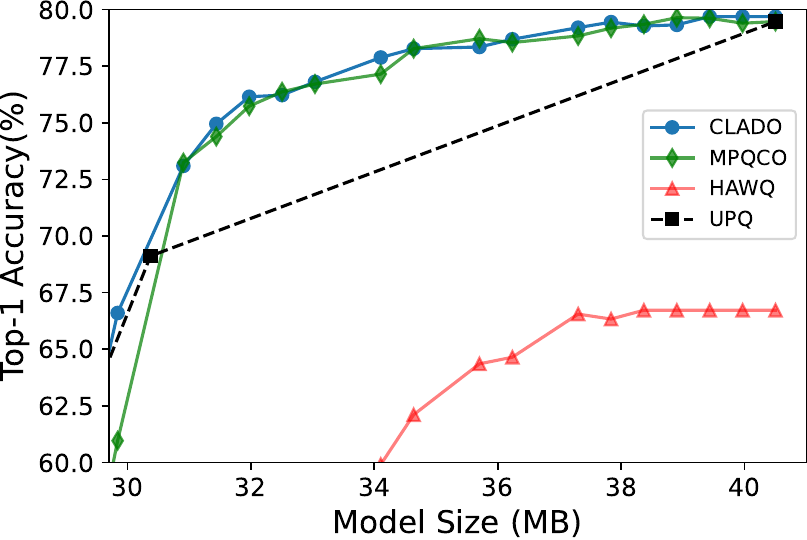}
        \caption{ViT-base}
        \label{fig:exp_vit}
    \end{subfigure}
    \caption{CNNs and ViT on the ImageNet dataset}
    \label{fig:exp_CNNs}
\end{figure}

\textbf{Implementation details:} All algorithms are implemented in PyTorch~\citep{Paszke2019PyTorchAI}.
The publicly available package CVXPY~\citep{Diamond2016CVXPYAP} is used to solve the IQP problem in CLADO with GUROBI~\citep{gurobi} as its backend. For fair comparisons, all algorithms are run using the same procedures and settings. We first download pretrained full-precision models from TorchVision. Then, model quantization is performed using MQBench~\citep{MQBench}.
Following MPQCO, quantization scale factors (and zero points in the affine case) are determined by minimization of the MSE between the \texttt{float32} values and their quantized values. After MQBench quantization, different MPQ algorithms are tested for mixed-precision quantization. 

\textbf{Use of multiple sensitivity sets:}
{Results of sensitivity-based algorithms depend on the samples in the sensitivity set and show some variation.}
To study the dependence of algorithms on the sensitivity set and to reduce the variation in results, we use multiple sensitivity sets: for each sample size ranging from 256 to 4096, we randomly construct 24 sets with the given size.
Then, we test the performance of algorithms on each of the 24 sets.

\subsection{Comparison of Different MPQ Algorithms}
\textbf{Runtime:}
CLADO requires $\frac{1}{2}|\mathbb{B}|I(|\mathbb{B}|I+1)$ measurements of the network output on the small sensitivity set. 
On a single Nvidia RTX2080 GPU, our implementation takes 1 hour to generate the sensitivities for the ResNet-34 and 2.5 hours for the ResNet-50 model. HAWQ takes roughly the same number of GPU-hours as CLADO to compute sensitivities. MPQCO is the fastest one, taking 5-10 minutes for its sensitivity estimation.

\begin{figure}[!t]
	\centering
	\begin{subfigure}{.45\textwidth}
		\includegraphics[width=\textwidth]{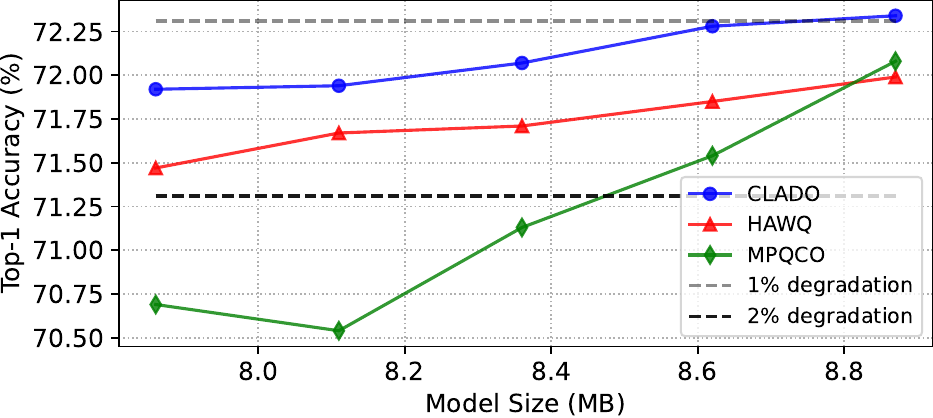}
		\caption{ResNet-34 (FP32 Accuracy 73.31\%)}
	\end{subfigure}
	\begin{subfigure}{0.45\textwidth}
            \includegraphics[width=\textwidth]{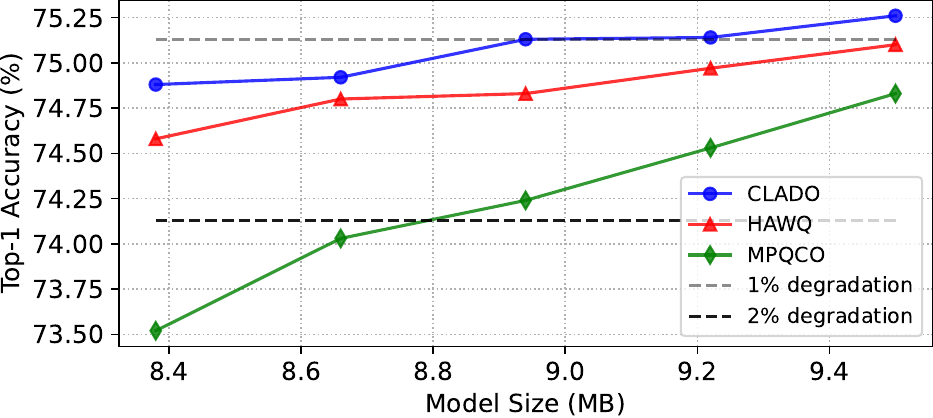}
    	\caption{ResNet-50 (FP32 Accuracy 76.13\%)}
	\end{subfigure}
    \caption{MPQ results (QAT): QAT fine-tuning based on CLADO outperforms fine-tuning based on other MPQ methods. (Range of model sizes is chosen to be close to 3-bit UPQ. With higher model size, all MPQ algorithms tend to recover FP32 performance.)}
    \label{fig:qat_mpq}
\end{figure}

\textbf{Average performance:}
Table \ref{tab:CLADO_PTQ_CNN} presents the averaged post-training quantization performance of algorithms under different model sizes (see Figure~\ref{fig:exp_CNNs} for tradeoff curves with more data points.)
CLADO delivers the best PTQ models under most constraints.
Its advantage is especially prominent when models are aggressively compressed to low precisions.
For example, under the smallest model size constraints in the table, CLADO demonstrates $5.73\%$ (ResNet-34) to $32\%$ (MobileNetV3) accuracy improvement over HAWQ and MPQCO. However, we note that the PTQ degradation in these low precision regions is too high and renders the results less meaningful. To address this, we apply additional quantization-aware fine-tuning. The QAT results are in Figure \ref{fig:qat_mpq}.

We found that, though CLADO is derived with (\ref{eq:FLDICT2}) under the PTQ settings, its bit-width assignments remain superior to others after fine-tuning. Since QAT significantly reduces degradation, the difference between algorithms' after-QAT performance is much smaller. E.g., for 8.87MB ResNet-34, CLADO, HAWQ, MPQCO achieve pre-QAT accuracies of $44.40\%$, $9.96\%$, and $0.51\%$, respectively, and post-QAT accuracies of $72.34\%$, $72.00\%$, and $72.08\%$.
\emph{Although the after-QAT difference is small, CLADO maintains meaningful advantage: e.g.}, for ResNet-34 with 8.87MB size constraint and ResNet-50 with 8.94MB, 9.22MB, and 9.50MB constraints, CLADO maintains $\le 1\%$ accuracy degradation while others exhibit a higher degradation.

\begin{figure}[!t]
\centering
    \begin{subfigure}{.45\textwidth}
        \includegraphics[width=\textwidth]{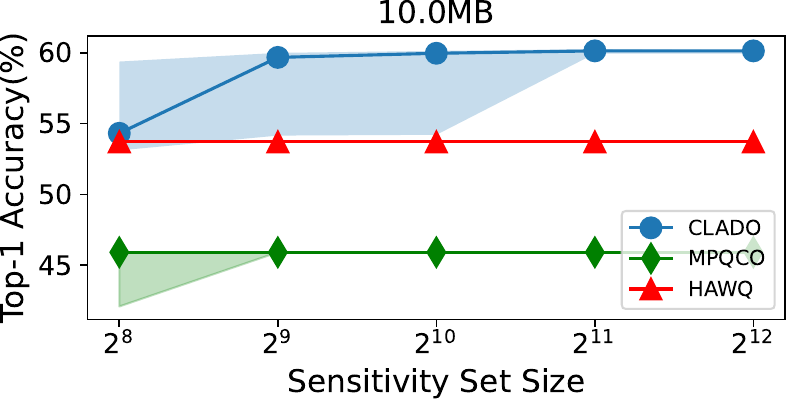}
    \end{subfigure} \hspace{3mm}
    \begin{subfigure}{.47\textwidth}
        \includegraphics[width=\textwidth]{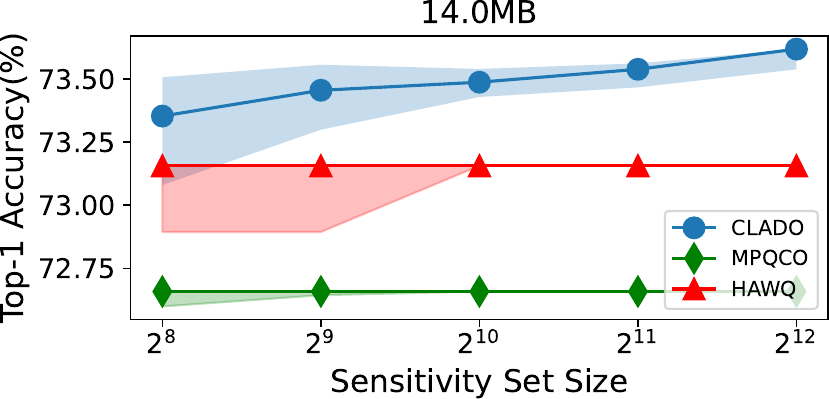}
    \end{subfigure}
 \caption{MPQ performance vs. sample size. Data shows median performance across 24 random sensitivity sets. Colored regions cover the upper and lower performance quartiles.}
 \label{fig:i1k-v}
\end{figure}

\textbf{Dependence on the sensitivity set:}
Figure \ref{fig:i1k-v} shows algorithms' variation due to different samples of the sensitivity set. For each sample size, it presents, the median, the upper quartile (75\%), and the lower quartile (25\%) of algorithms' performance on the 24 randomly sampled sensitivity sets. CLADO produces reasonably stable results. Though it may have slightly higher variation sometimes, its lower quartile performance is almost always better than alternatives' upper quartile performance, which is especially true when the sample size is $\geq 1024$ on ImageNet.

\begin{figure*}[]
    \centering
    \begin{subfigure}{0.99\textwidth}
        \includegraphics[width=.95\textwidth]{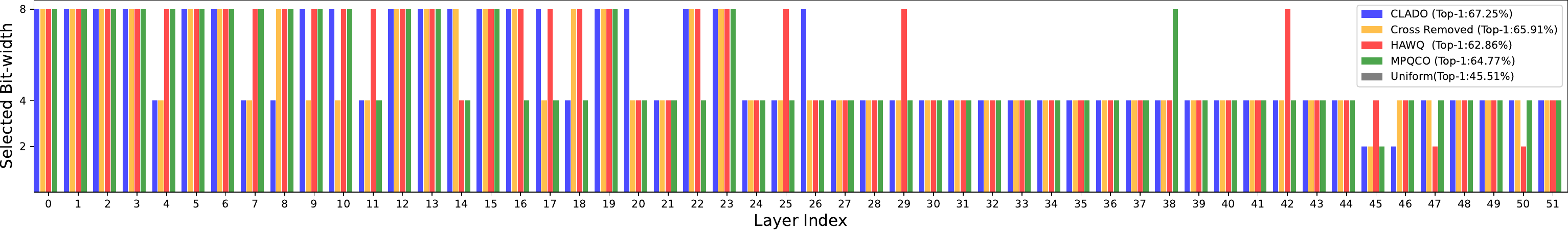}
    \end{subfigure}

\scriptsize
\setlength{\tabcolsep}{1pt}
\begin{tabular}{|llllllllllllllllllllllllllllllllllllllllllllllllllll|}
\hline
\rotatebox[origin=c]{90}{0} & \rotatebox[origin=c]{90}{1} & \rotatebox[origin=c]{90}{2} & \rotatebox[origin=c]{90}{3} & \rotatebox[origin=c]{90}{4} & \rotatebox[origin=c]{90}{5} & \rotatebox[origin=c]{90}{6} & \rotatebox[origin=c]{90}{7} & \rotatebox[origin=c]{90}{8} & \rotatebox[origin=c]{90}{9} & \rotatebox[origin=c]{90}{10} & \rotatebox[origin=c]{90}{11} & \rotatebox[origin=c]{90}{12} & \rotatebox[origin=c]{90}{13} & \rotatebox[origin=c]{90}{14} & \rotatebox[origin=c]{90}{15} & \rotatebox[origin=c]{90}{16} & \rotatebox[origin=c]{90}{17} & \rotatebox[origin=c]{90}{18} & \rotatebox[origin=c]{90}{19} & \rotatebox[origin=c]{90}{20} & \rotatebox[origin=c]{90}{21} & \rotatebox[origin=c]{90}{22} & \rotatebox[origin=c]{90}{23} & \rotatebox[origin=c]{90}{24} & \rotatebox[origin=c]{90}{25} & \rotatebox[origin=c]{90}{26} & \rotatebox[origin=c]{90}{27} & \rotatebox[origin=c]{90}{28} & \rotatebox[origin=c]{90}{29} & \rotatebox[origin=c]{90}{30} & \rotatebox[origin=c]{90}{31} & \rotatebox[origin=c]{90}{32} & \rotatebox[origin=c]{90}{33} & \rotatebox[origin=c]{90}{34} & \rotatebox[origin=c]{90}{35} & \rotatebox[origin=c]{90}{36} & \rotatebox[origin=c]{90}{37} & \rotatebox[origin=c]{90}{38} & \rotatebox[origin=c]{90}{39} & \rotatebox[origin=c]{90}{40} & \rotatebox[origin=c]{90}{41} & \rotatebox[origin=c]{90}{42} & \rotatebox[origin=c]{90}{43} & \rotatebox[origin=c]{90}{44} & \rotatebox[origin=c]{90}{45} & \rotatebox[origin=c]{90}{46} & \rotatebox[origin=c]{90}{47} & \rotatebox[origin=c]{90}{48} & \rotatebox[origin=c]{90}{49} & \rotatebox[origin=c]{90}{50} & \rotatebox[origin=c]{90}{51} \\ \hline
\rotatebox[origin=c]{90}{layer1.0.conv1} & \rotatebox[origin=c]{90}{layer1.0.conv2} & \rotatebox[origin=c]{90}{layer1.0.conv3} & \rotatebox[origin=c]{90}{layer1.0.downsample.0} & \rotatebox[origin=c]{90}{layer1.1.conv1} & \rotatebox[origin=c]{90}{layer1.1.conv2} & \rotatebox[origin=c]{90}{layer1.1.conv3} & \rotatebox[origin=c]{90}{layer1.2.conv1} & \rotatebox[origin=c]{90}{layer1.2.conv2} & \rotatebox[origin=c]{90}{layer1.2.conv3} & \rotatebox[origin=c]{90}{layer2.0.conv1} & \rotatebox[origin=c]{90}{layer2.0.conv2} & \rotatebox[origin=c]{90}{layer2.0.conv3} & \rotatebox[origin=c]{90}{layer2.0.downsample.0} & \rotatebox[origin=c]{90}{layer2.1.conv1} & \rotatebox[origin=c]{90}{layer2.1.conv2} & \rotatebox[origin=c]{90}{layer2.1.conv3} & \rotatebox[origin=c]{90}{layer2.2.conv1} & \rotatebox[origin=c]{90}{layer2.2.conv2} & \rotatebox[origin=c]{90}{layer2.2.conv3} & \rotatebox[origin=c]{90}{layer2.3.conv1} & \rotatebox[origin=c]{90}{layer2.3.conv2} & \rotatebox[origin=c]{90}{layer2.3.conv3} & \rotatebox[origin=c]{90}{layer3.0.conv1} & \rotatebox[origin=c]{90}{layer3.0.conv2} & \rotatebox[origin=c]{90}{layer3.0.conv3} & \rotatebox[origin=c]{90}{layer3.0.downsample.0} & \rotatebox[origin=c]{90}{layer3.1.conv1} & \rotatebox[origin=c]{90}{layer3.1.conv2} & \rotatebox[origin=c]{90}{layer3.1.conv3} & \rotatebox[origin=c]{90}{layer3.2.conv1} & \rotatebox[origin=c]{90}{layer3.2.conv2} & \rotatebox[origin=c]{90}{layer3.2.conv3} & \rotatebox[origin=c]{90}{layer3.3.conv1} & \rotatebox[origin=c]{90}{layer3.3.conv2} & \rotatebox[origin=c]{90}{layer3.3.conv3} & \rotatebox[origin=c]{90}{layer3.4.conv1} & \rotatebox[origin=c]{90}{layer3.4.conv2} & \rotatebox[origin=c]{90}{layer3.4.conv3} & \rotatebox[origin=c]{90}{layer3.5.conv1} & \rotatebox[origin=c]{90}{layer3.5.conv2} & \rotatebox[origin=c]{90}{layer3.5.conv3} & \rotatebox[origin=c]{90}{layer4.0.conv1} & \rotatebox[origin=c]{90}{layer4.0.conv2} & \rotatebox[origin=c]{90}{layer4.0.conv3} & \rotatebox[origin=c]{90}{layer4.0.downsample.0} & \rotatebox[origin=c]{90}{layer4.1.conv1} & \rotatebox[origin=c]{90}{layer4.1.conv2} & \rotatebox[origin=c]{90}{layer4.1.conv3} & \rotatebox[origin=c]{90}{layer4.2.conv1} & \rotatebox[origin=c]{90}{layer4.2.conv2} & \rotatebox[origin=c]{90}{layer4.2.conv3}\\
\hline

\end{tabular}
    \caption{Bit-width assignments to ResNet-50 along with the layer index-layer name mappings. Model size constraint is set to be 11.18MB, corresponding to 4-bit UPQ.}
    \label{fig:decisions-r50}
\end{figure*}

\textbf{Analysis of bit-width assignments:}
Figure~\ref{fig:decisions-r50} visualizes the MPQ decisions using different algorithms on ResNet-50. Different MPQ algorithms share some commonalities, \emph{e.g.}, generally they assign more bits to shallow layers and less bits to deep layers.
This aligns with our observation that shallow layers have higher sensitivities and fewer parameters. However, algorithms decide quite differently for certain layers: compared to HAWQ and MPQCO, CLADO chooses to quantize more aggressively on the first convolutional layer of \texttt{layer1.1} (index 4), and \texttt{layer4.1} (index 46) and more conservatively on the downsampling layer of \texttt{layer3.0} (index 26). Additional results on bit-width assignment for various networks and model size constraints are provided in Section~\ref{sec:app_bitwidth} of the Appendix.

\section{Experimental Results: Vision Transformer}
\label{sec:exp_transformer}
We also tested our MPQ flow on a transformer-based vision model. The Vision Transformer (ViT) has significantly advanced computer vision by applying transformer architectures, originally designed for natural language processing, to visual data~\citep{dosovitskiy2020image}. Its ability to capture long-range dependencies and global context has shown superior performance compared to traditional convolutional neural networks (CNNs) in tasks like image classification and object detection. ViT is also commonly used in CLIP~\citep{radford2021learning} as an image encoder to extract visual features, enabling zero-shot image classification and connecting vision to text. We use a pretrained ViT-based model from the HuggingFace library~\citep{wolf-etal-2020-transformers} and evaluate the quantized model on the ImageNet dataset using various PTQ algorithms~\citep{Russakovsky2014ImageNetLS}.

Table~\ref{tab:CLADO_PTQ_CNN} shows the size/performance tradeoffs (see Figure~\ref{fig:exp_vit} for full Pareto fronts). We found that \emph{CLADO almost always produces the best MPQ solutions.} The advantage of CLADO over the next best alternative is small under large size constraints. Under more restrictive size constraints, CLADO becomes more advantageous.

\section{Ablation Studies}
\label{sec:ablation}

\begin{figure*}[!t]
    \centering
    \begin{subfigure}{.22\textwidth}
        \includegraphics[width=.99\textwidth]{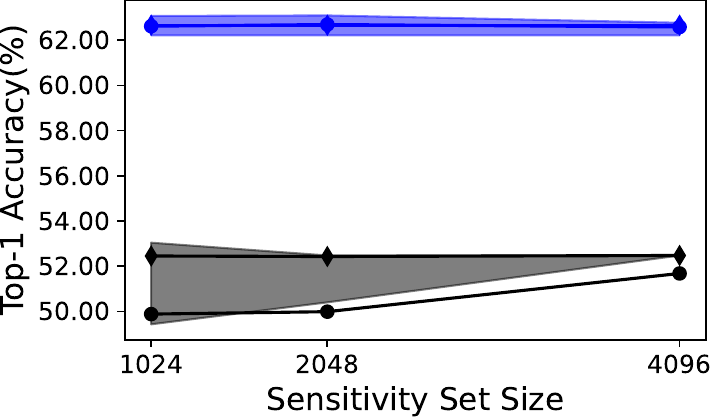}
        \caption{ResNet-34: 10.14MB}
    \end{subfigure} 
    \begin{subfigure}{.21\textwidth}
        \includegraphics[width=.99\textwidth]{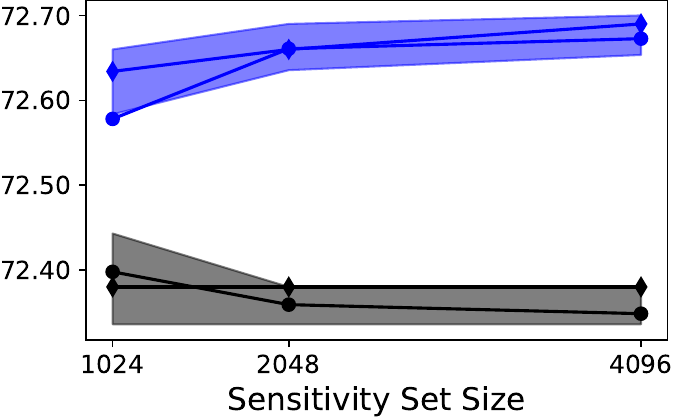}
        \caption{ResNet-34: 16.22MB}
    \end{subfigure} 
    \begin{subfigure}{.21\textwidth}
        \includegraphics[width=.99\textwidth]{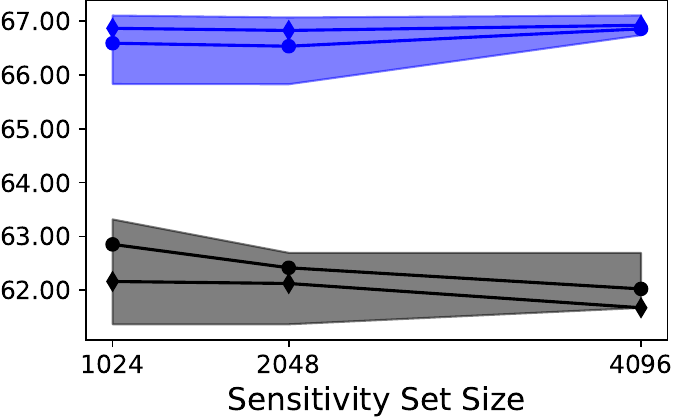}
        \caption{ResNet-50: 11.18MB}
    \end{subfigure}  
    \begin{subfigure}{.31\textwidth}
        \includegraphics[width=.99\textwidth]{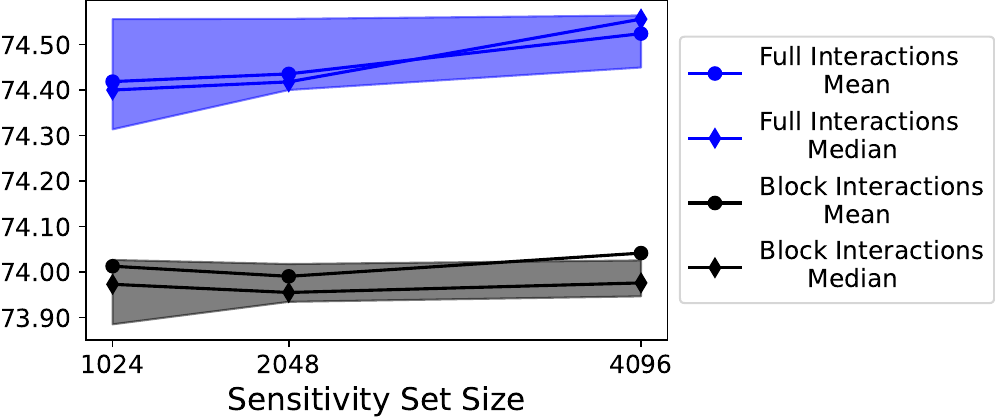}
        \caption{ResNet-50: 15.65MB}
    \end{subfigure}
    \caption{Leaving out inter-block dependencies  worsens MPQ. 
   Blue: CLADO, captures all-layer dependencies. Black: inter-block dependencies ignored, following BRECQ~\citep{Li2021BRECQPT} Data represents median performance on 24 randomly sampled sensitivity sets. Colored regions cover the upper and lower quartiles of performance.}
    \label{fig:FIvsBI}
\end{figure*}

\textbf{Cross-Layer dependencies:} To demonstrate the importance of cross-layer dependencies, we test the performance of CLADO assuming zero cross-layer sensitivities for all layers. We refer to this set of experiments as ``CLADO*''. Compared to full CLADO, \emph{ignoring cross-layer dependencies consistently and significantly degrades MPQ performance} (Table~\ref{tab:CLADO_PTQ_CNN}). Cross-layer dependencies seem to be rather prominent for ViT and for CNN compression under small size constraints. 

BRECQ~\citep{Li2021BRECQPT} claimed that block-level optimization performs best for quantization by adaptive rounding. 
Interestingly, however, we show that this observation does not extend to MPQ. Following BRECQ, we conducted ablation experiments in which cross-layer dependencies are only considered within blocks of layers. In this set of experiments, we intentionally leave out the inter-block dependencies. We refer to this group of experiments as ``block interactions'' (the black curves in Figure~\ref{fig:FIvsBI}). 
\emph{We found that ignoring inter-block dependencies significantly worsens the MPQ solution.} We hypothesize that MPQ, with far fewer parameters ($|\mathbb{B}|I$) than adaptive rounding, is less prone to the overfitting issue noted by BRECQ. Thus, using full pairwise interactions yields more robust results, while limiting to intra-block interactions likely underfits MPQ solutions.

\begin{figure*}[!t]
    \centering
    \begin{subfigure}{.22\textwidth}
        \includegraphics[width=.99\textwidth]{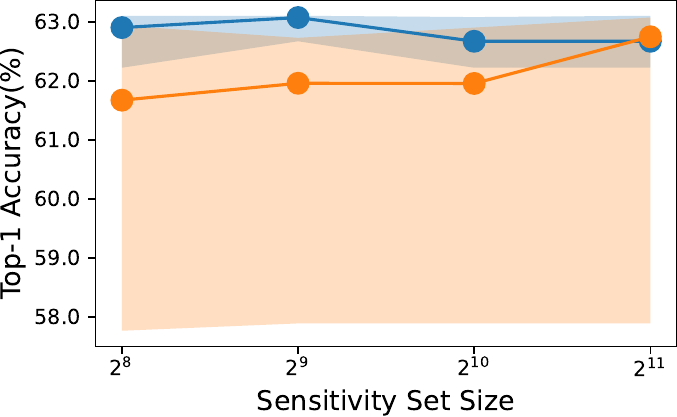}
        \caption{ResNet-34, 10.1MB}
    \end{subfigure}
    \begin{subfigure}{.21\textwidth}
        \includegraphics[width=.99\textwidth]{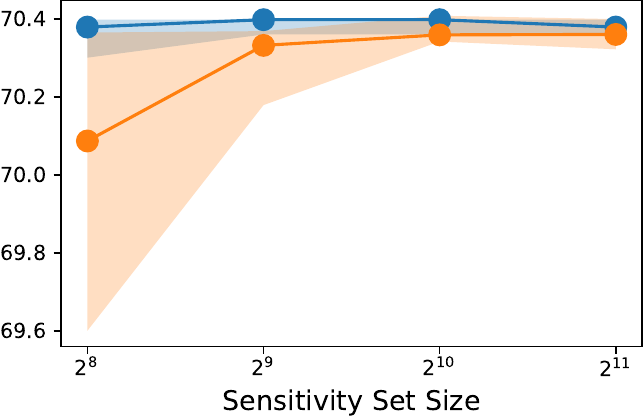}
        \caption{ResNet-34, 12.2MB}
    \end{subfigure}
    \begin{subfigure}{.21\textwidth}
        \includegraphics[width=.99\textwidth]{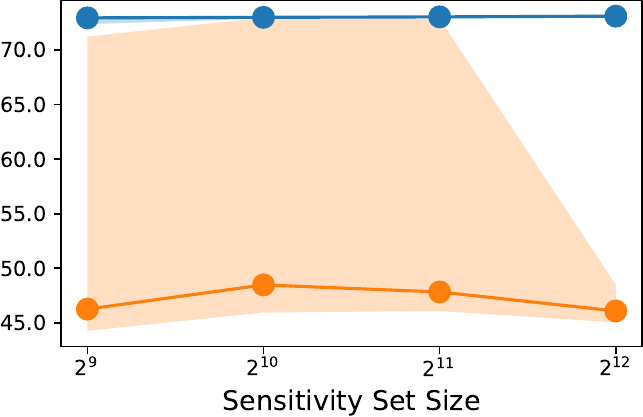}
        \caption{ResNet-50, 13.4MB}
    \end{subfigure}
    \begin{subfigure}{.29\textwidth}
        \includegraphics[width=.99\textwidth]{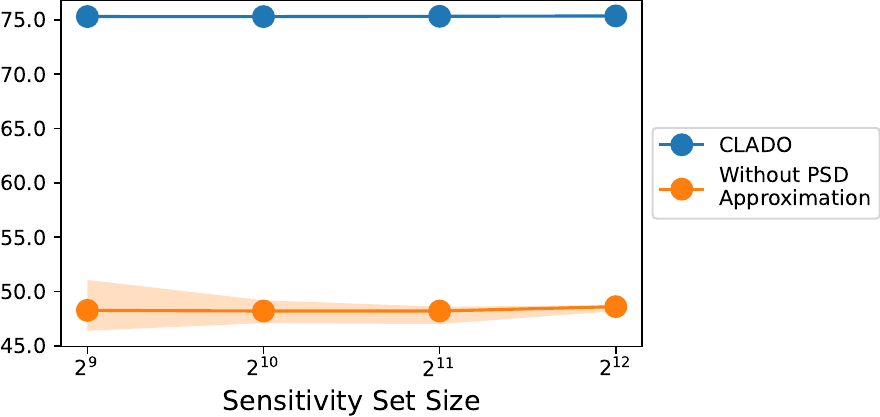}
        \caption{ResNet-50, 17.9MB}
    \end{subfigure}
    \caption{Ablation on PSD Approximation of the Sensitivity Matrix $\hat{G}$.}
    \label{fig:i1k-abl-PSD}
\end{figure*}

\begin{table*}[!t]
\small
\centering
    \caption{Approx. accuracy of the second-order quantization error ($vHv$): our fast method vs. the exact Hessian evaluation.}
    \resizebox{0.98\textwidth}{!}{
        \begin{tabular}{l|c|c|c|c|c|c|c}
        \toprule
        \begin{tabular}[c]{@{}l@{}}ResNet20\\ Layers\end{tabular} & \begin{tabular}[c]{@{}l@{}}module.layer1\\ .0.conv1 (2-bit)\end{tabular}  & \begin{tabular}[c]{@{}l@{}}module.layer2\\ .1.conv2 (6-bit)\end{tabular} & \begin{tabular}[c]{@{}l@{}}module.layer3\\ .2.conv1 (2-bit)\end{tabular} & \begin{tabular}[c]{@{}l@{}}module.layer3\\ .2.conv1 (4-bit)\end{tabular} & \begin{tabular}[c]{@{}l@{}}module.layer3\\ .2.conv1 (6-bit)\end{tabular} & \begin{tabular}[c]{@{}l@{}}module.fc\\ (2-bit)\end{tabular} & \begin{tabular}[c]{@{}l@{}}module.fc\\ (4-bit)\end{tabular}  \\ \hline
        vHv (exact) & 0.17105  & 0.00005 & 0.09384 & 0.00374  & 0.00037  & 0.03154  & 0.00057  \\
        vHv (ours)  & 0.14670  & 0.00006 & 0.09912 & 0.00362  & 0.00038  & 0.03190  & 0.00057  \\
        \bottomrule
        \end{tabular}
    }
    \label{tab:hessian_acc}
\end{table*}

\textbf{PSD sensitivity matrix approximation:}
PSD approximation is critical to CLADO's superior performance over other studied algorithms. To explore this, we disable the use of PSD approximation in CLADO, keeping the rest of CLADO unchanged. We found that it makes CVXPY unable to converge on a solution in more than 3 hours (with the PSD approximation, the solver computes the optimal solutions within seconds). This observation aligns with our expectation that the convexity of the objective $\Omega$ makes the IQP more well-behaved and easier to solve. The results are captured by the purple curves in Figures~\ref{fig:i1k-abl-PSD}: though PSD does not always produce best results, it improves their consistency, and, in some cases, prevents a severe degradation in the quality of solutions.

\textbf{Approximation accuracy of our fast Hessian-free method:} We evaluate the accuracy of our fast method that avoids the Hessian computation by comparing the quantization-induced loss changes with those from the exact Hessian method, which is $7\times$ slower and requires more CUDA memory. Results for randomly selected shallow and deep layers in ResNet20 are shown in Table~\ref{tab:hessian_acc}, where $v$ represents the quantization error at each bit-width. The approximation error is generally small.

\section{Conclusions}
\label{sec:conclusions}
We present CLADO, a sensitivity-based mixed-precision quantization algorithm that exploits cross-layer quantization error dependencies. We derive a sensitivity objective from $2^{nd}$-order Taylor expansion of loss, and reformulate the bit-width assignment problem as an integer quadratic program that can be solved efficiently. We devise an efficient algorithm to compute the cross-layer dependencies that requires only forward evaluations of networks. Extensive experiments are conducted to demonstrate the effectiveness of the proposed method.

\bibliographystyle{abbrvnat}
\bibliography{templateArxiv}

\begin{thebibliography}{32}
\providecommand{\natexlab}[1]{#1}
\providecommand{\url}[1]{\texttt{#1}}
\expandafter\ifx\csname urlstyle\endcsname\relax
  \providecommand{\doi}[1]{doi: #1}\else
  \providecommand{\doi}{doi: \begingroup \urlstyle{rm}\Url}\fi

\bibitem[Azizi et~al.(2024)Azizi, Nazemi, Fayyazi, and Pedram]{azizi2024automated}
S.~Azizi, M.~Nazemi, A.~Fayyazi, and M.~Pedram.
\newblock Automated optimization of deep neural networks: Dynamic bit-width and layer-width selection via cluster-based parzen estimation.
\newblock In \emph{Design, Automation \& Test in Europe Conference \& Exhibition (DATE)}, 2024.

\bibitem[Cai et~al.(2020)Cai, Yao, Dong, Gholami, Mahoney, and Keutzer]{Cai2020ZeroQAN}
Y.~Cai, Z.~Yao, Z.~Dong, A.~Gholami, M.~W. Mahoney, and K.~Keutzer.
\newblock {ZeroQ}: A novel zero shot quantization framework.
\newblock \emph{IEEE/CVF Conference on Computer Vision and Pattern Recognition (CVPR)}, pages 13166--13175, 2020.

\bibitem[Chauhan et~al.(2023)Chauhan, Tiwari, and R]{Chauhan2023PostTM}
A.~Chauhan, U.~Tiwari, and V.~N. R.
\newblock Post training mixed precision quantization of neural networks using first-order information.
\newblock \emph{IEEE/CVF International Conference on Computer Vision Workshops (ICCVW)}, pages 1335--1344, 2023.

\bibitem[Chen et~al.(2021)Chen, Wang, and Cheng]{Chen2021TowardsMQ}
W.~Chen, P.~Wang, and J.~Cheng.
\newblock Towards mixed-precision quantization of neural networks via constrained optimization.
\newblock \emph{IEEE/CVF International Conference on Computer Vision (ICCV)}, pages 5330--5339, 2021.

\bibitem[Chu et~al.(2024)Chu, Li, and Zhang]{Chu_Li_Zhang_2024}
X.~Chu, L.~Li, and B.~Zhang.
\newblock Make repvgg greater again: A quantization-aware approach.
\newblock \emph{Proceedings of the AAAI Conference on Artificial Intelligence}, 38:\penalty0 11624--11632, 2024.

\bibitem[Courbariaux et~al.(2015)Courbariaux, Bengio, and David]{Courbariaux2015BinaryConnectTD}
M.~Courbariaux, Y.~Bengio, and J.-P. David.
\newblock {BinaryConnect}: Training deep neural networks with binary weights during propagations.
\newblock \emph{Advances in neural information processing systems}, 28, 2015.

\bibitem[Diamond and Boyd(2016)]{Diamond2016CVXPYAP}
S.~Diamond and S.~P. Boyd.
\newblock Cvxpy: A python-embedded modeling language for convex optimization.
\newblock \emph{Journal of machine learning research : JMLR}, 17, 2016.

\bibitem[Dong et~al.(2023)Dong, Li, Wei, Niu, Tian, and Pan]{Dong2023EMQET}
P.~Dong, L.~Li, Z.~Wei, X.-Y. Niu, Z.~Tian, and H.~Pan.
\newblock {EMQ}: Evolving training-free proxies for automated mixed precision quantization.
\newblock \emph{IEEE/CVF International Conference on Computer Vision (ICCV)}, pages 17030--17040, 2023.

\bibitem[Dong et~al.(2019)Dong, Yao, Gholami, Mahoney, and Keutzer]{Dong2019HAWQHA}
Z.~Dong, Z.~Yao, A.~Gholami, M.~W. Mahoney, and K.~Keutzer.
\newblock {HAWQ}: Hessian aware quantization of neural networks with mixed-precision.
\newblock \emph{IEEE/CVF International Conference on Computer Vision (ICCV)}, pages 293--302, 2019.

\bibitem[Dong et~al.(2020)Dong, Yao, Cai, Arfeen, Gholami, Mahoney, and Keutzer]{Dong2020HAWQV2HA}
Z.~Dong, Z.~Yao, Y.~Cai, D.~Arfeen, A.~Gholami, M.~W. Mahoney, and K.~Keutzer.
\newblock {HAWQ-V2}: Hessian aware trace-weighted quantization of neural networks.
\newblock \emph{ArXiv}, abs/1911.03852, 2020.

\bibitem[Dosovitskiy et~al.(2020)Dosovitskiy, Beyer, Kolesnikov, Weissenborn, Zhai, Unterthiner, Dehghani, Minderer, Heigold, Gelly, Uszkoreit, and Houlsby]{dosovitskiy2020image}
A.~Dosovitskiy, L.~Beyer, A.~Kolesnikov, D.~Weissenborn, X.~Zhai, T.~Unterthiner, M.~Dehghani, M.~Minderer, G.~Heigold, S.~Gelly, J.~Uszkoreit, and N.~Houlsby.
\newblock An image is worth 16x16 words: Transformers for image recognition at scale.
\newblock \emph{ArXiv}, abs/2010.11929, 2020.

\bibitem[Guo et~al.(2020)Guo, Zhang, Mu, Heng, Liu, Wei, and Sun]{Guo2020SinglePO}
Z.~Guo, X.~Zhang, H.~Mu, W.~Heng, Z.~Liu, Y.~Wei, and J.~Sun.
\newblock Single path one-shot neural architecture search with uniform sampling.
\newblock In \emph{ECCV}, 2020.

\bibitem[{Gurobi Optimization, LLC}(2023)]{gurobi}
{Gurobi Optimization, LLC}.
\newblock {Gurobi Optimizer Reference Manual}, 2023.
\newblock URL \url{https://www.gurobi.com}.

\bibitem[He et~al.(2015)He, Zhang, Ren, and Sun]{He2015DeepRL}
K.~He, X.~Zhang, S.~Ren, and J.~Sun.
\newblock Deep residual learning for image recognition.
\newblock \emph{IEEE/CVF Conference on Computer Vision and Pattern Recognition (CVPR)}, pages 770--778, 2015.

\bibitem[Howard et~al.(2019)Howard, Sandler, Chu, Chen, Chen, Tan, Wang, Zhu, Pang, Vasudevan, Le, and Adam]{Howard2019SearchingFM}
A.~G. Howard, M.~Sandler, G.~Chu, L.-C. Chen, B.~Chen, M.~Tan, W.~Wang, Y.~Zhu, R.~Pang, V.~Vasudevan, Q.~V. Le, and H.~Adam.
\newblock Searching for mobilenetv3.
\newblock \emph{IEEE/CVF International Conference on Computer Vision (ICCV)}, pages 1314--1324, 2019.

\bibitem[Kim et~al.(2021)Kim, Gholami, Yao, Mahoney, and Keutzer]{Kim2021IBERTIB}
S.~Kim, A.~Gholami, Z.~Yao, M.~W. Mahoney, and K.~Keutzer.
\newblock Integer-only bert quantization.
\newblock \emph{ArXiv}, abs/2101.01321, 2021.

\bibitem[Kundu et~al.(2022)Kundu, Wang, Sun, Beerel, and Pedram]{kundu2022bmpq}
S.~Kundu, S.~Wang, Q.~Sun, P.~A. Beerel, and M.~Pedram.
\newblock {BMPQ}: bit-gradient sensitivity-driven mixed-precision quantization of dnns from scratch.
\newblock In \emph{Design, Automation \& Test in Europe Conference \& Exhibition (DATE)}, pages 588--591, 2022.

\bibitem[Li et~al.(2021)Li, Gong, Tan, Yang, Hu, Zhang, Yu, Wang, and Gu]{Li2021BRECQPT}
Y.~Li, R.~Gong, X.~Tan, Y.~Yang, P.~Hu, Q.~Zhang, F.~Yu, W.~Wang, and S.~Gu.
\newblock {BRECQ}: Pushing the limit of post-training quantization by block reconstruction.
\newblock \emph{ArXiv}, abs/2102.05426, 2021.

\bibitem[Li* et~al.(2021)Li*, Shen*, Ma*, Ren*, Zhao*, Zhang*, Gong*, Yu, and Yan]{MQBench}
Y.~Li*, M.~Shen*, J.~Ma*, Y.~Ren*, M.~Zhao*, Q.~Zhang*, R.~Gong*, F.~Yu, and J.~Yan.
\newblock Mqbench: Towards reproducible and deployable model quantization benchmark.
\newblock \emph{Proceedings of the Neural Information Processing Systems Track on Datasets and Benchmarks}, 2021.

\bibitem[Lou et~al.(2020)Lou, Guo, Liu, Kim, and Jiang]{Lou2020AutoQAK}
Q.~Lou, F.~Guo, L.~Liu, M.~Kim, and L.~Jiang.
\newblock {AutoQ}: Automated kernel-wise neural network quantization.
\newblock \emph{ArXiv}, abs/1902.05690, 2020.

\bibitem[Ma et~al.(2021)Ma, Jin, Zheng, Wang, Li, Jiang, Zhang, and Ji]{Ma2021OMPQOM}
Y.~Ma, T.~Jin, X.~Zheng, Y.~Wang, H.~Li, G.~Jiang, W.~Zhang, and R.~Ji.
\newblock {OMPQ}: Orthogonal mixed precision quantization.
\newblock \emph{ArXiv}, abs/2109.07865, 2021.

\bibitem[Nagel et~al.(2020)Nagel, Amjad, van Baalen, Louizos, and Blankevoort]{Nagel2020UpOD}
M.~Nagel, R.~A. Amjad, M.~van Baalen, C.~Louizos, and T.~Blankevoort.
\newblock Up or down? adaptive rounding for post-training quantization.
\newblock \emph{ArXiv}, abs/2004.10568, 2020.

\bibitem[Paszke et~al.(2019)Paszke, Gross, Massa, Lerer, Bradbury, Chanan, Killeen, Lin, Gimelshein, Antiga, et~al.]{Paszke2019PyTorchAI}
A.~Paszke, S.~Gross, F.~Massa, A.~Lerer, J.~Bradbury, G.~Chanan, T.~Killeen, Z.~Lin, N.~Gimelshein, L.~Antiga, et~al.
\newblock {PyTorch}: An imperative style, high-performance deep learning library.
\newblock \emph{Advances in neural information processing systems}, 32, 2019.

\bibitem[Radford et~al.(2021)Radford, Kim, Hallacy, Ramesh, Goh, Agarwal, Sastry, Askell, Mishkin, Clark, et~al.]{radford2021learning}
A.~Radford, J.~W. Kim, C.~Hallacy, A.~Ramesh, G.~Goh, S.~Agarwal, G.~Sastry, A.~Askell, P.~Mishkin, J.~Clark, et~al.
\newblock Learning transferable visual models from natural language supervision.
\newblock In \emph{International conference on machine learning}, pages 8748--8763. PMLR, 2021.

\bibitem[Radosavovic et~al.(2020)Radosavovic, Kosaraju, Girshick, He, and Doll{\'a}r]{Radosavovic2020DesigningND}
I.~Radosavovic, R.~P. Kosaraju, R.~Girshick, K.~He, and P.~Doll{\'a}r.
\newblock Designing network design spaces.
\newblock \emph{IEEE/CVF Conference on Computer Vision and Pattern Recognition (CVPR)}, pages 10425--10433, 2020.

\bibitem[Russakovsky et~al.(2014)Russakovsky, Deng, Su, Krause, Satheesh, Ma, Huang, Karpathy, Khosla, Bernstein, et~al.]{Russakovsky2014ImageNetLS}
O.~Russakovsky, J.~Deng, H.~Su, J.~Krause, S.~Satheesh, S.~Ma, Z.~Huang, A.~Karpathy, A.~Khosla, M.~Bernstein, et~al.
\newblock Imagenet large scale visual recognition challenge.
\newblock \emph{International Journal of Computer Vision}, 115:\penalty0 211--252, 2014.

\bibitem[Tang et~al.(2022)Tang, Ouyang, Wang, Zhu, Wang, Ji, and Zhu]{Tang2022MixedPrecisionNN}
C.~Tang, K.~Ouyang, Z.~Wang, Y.~Zhu, Y.~Wang, W.~Ji, and W.~Zhu.
\newblock Mixed-precision neural network quantization via learned layer-wise importance.
\newblock \emph{ArXiv}, abs/2203.08368, 2022.

\bibitem[Wang et~al.(2019)Wang, Liu, Lin, Lin, and Han]{Wang2019HAQHA}
K.~Wang, Z.~Liu, Y.~Lin, J.~Lin, and S.~Han.
\newblock {HAQ}: Hardware-aware automated quantization with mixed precision.
\newblock \emph{IEEE/CVF Conference on Computer Vision and Pattern Recognition (CVPR)}, pages 8604--8612, 2019.

\bibitem[Wei et~al.(2022)Wei, Gong, Li, Liu, and Yu]{Wei2022QDropRD}
X.~Wei, R.~Gong, Y.~Li, X.~Liu, and F.~Yu.
\newblock {QDrop}: Randomly dropping quantization for extremely low-bit post-training quantization.
\newblock \emph{ArXiv}, abs/2203.05740, 2022.

\bibitem[Wolf et~al.(2020)Wolf, Debut, Sanh, Chaumond, Delangue, Moi, Cistac, Rault, Louf, Funtowicz, et~al.]{wolf-etal-2020-transformers}
T.~Wolf, L.~Debut, V.~Sanh, J.~Chaumond, C.~Delangue, A.~Moi, P.~Cistac, T.~Rault, R.~Louf, M.~Funtowicz, et~al.
\newblock Transformers: State-of-the-art natural language processing.
\newblock In \emph{Proceedings of the conference on empirical methods in natural language processing: system demonstrations}, pages 38--45, 2020.

\bibitem[Wu et~al.(2018)Wu, Wang, Zhang, Tian, Vajda, and Keutzer]{Wu2018MixedPQ}
B.~Wu, Y.~Wang, P.~Zhang, Y.~Tian, P.~Vajda, and K.~Keutzer.
\newblock Mixed precision quantization of convnets via differentiable neural architecture search.
\newblock \emph{ArXiv}, abs/1812.00090, 2018.

\bibitem[Yao et~al.(2021)Yao, Dong, Zheng, Gholami, Yu, Tan, Wang, Huang, Wang, Mahoney, et~al.]{yao2021hawq}
Z.~Yao, Z.~Dong, Z.~Zheng, A.~Gholami, J.~Yu, E.~Tan, L.~Wang, Q.~Huang, Y.~Wang, M.~Mahoney, et~al.
\newblock {HAWQ-V3}: Dyadic neural network quantization.
\newblock In \emph{International Conference on Machine Learning}, pages 11875--11886. PMLR, 2021.

\end{thebibliography}

\newpage
\appendix
\section{Indexing Layers in Different Models}

\begin{figure}[!htb]
    \centering
        \begin{subfigure}{.2\textwidth}
                \centering
        	\includegraphics[width=\textwidth]{figs/r34_layerindex.pdf}
                \caption{ResNet-34}
        \end{subfigure}
        \begin{subfigure}{.2\textwidth}
                \centering
        	\includegraphics[width=\textwidth]{figs/r50_layerindex.pdf}
                \caption{ResNet-50}
        \end{subfigure}
        \begin{subfigure}{.2\textwidth}
                \centering
        	\includegraphics[width=\textwidth]{figs/mbnv3_layerindex.pdf}
                \caption{MobileNetV3}
        \end{subfigure}
        \begin{subfigure}{.27\textwidth}
                \centering
        	\includegraphics[width=\textwidth]{figs/vit_layerindex.pdf}
                \caption{ViT}
        \end{subfigure}
     \caption{Layer indices}
     \label{fig:layerindex}
    \end{figure}

\newpage
\section{Bit-width Assignment Visualizations}
\label{sec:app_bitwidth}
\begin{figure}[!htb]
    \centering
        \begin{subfigure}{\textwidth}
                \centering
        	\includegraphics[width=\textwidth]{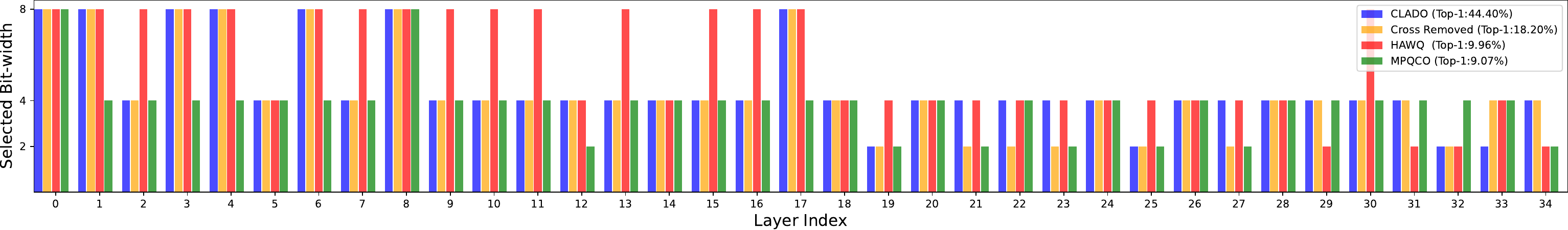}
                \caption{8.87MB}
        \end{subfigure}
        \begin{subfigure}{\textwidth}
                \centering
        	\includegraphics[width=\textwidth]{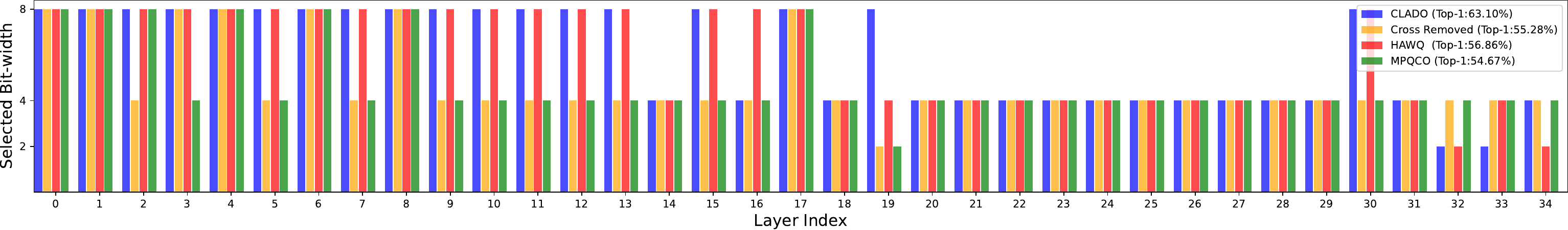}
                \caption{10.17MB}
        \end{subfigure}
     \caption{ResNet-34}
     \label{fig:r34dec}
\end{figure}

\begin{figure}[!htb]
    \centering
        \begin{subfigure}{\textwidth}
                \centering
        	\includegraphics[width=\textwidth]{figs/decsions_r50_11.18MB_naive_simplex.pdf}
                \caption{11.18MB}
        \end{subfigure}
        \begin{subfigure}{\textwidth}
                \centering
        	\includegraphics[width=\textwidth]{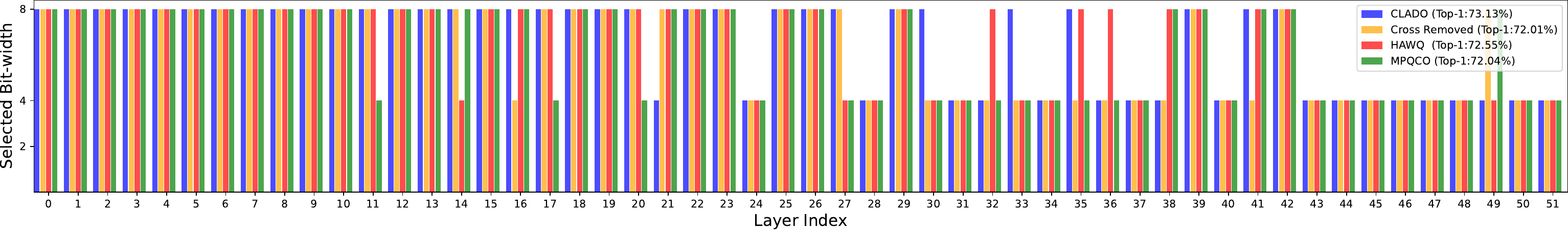}
                \caption{13.41MB}
        \end{subfigure}
     \caption{ResNet-50}
     \label{fig:r50dec}
\end{figure}

\begin{figure}[!htb]
    \centering
        \begin{subfigure}{\textwidth}
                \centering
        	\includegraphics[width=\textwidth]{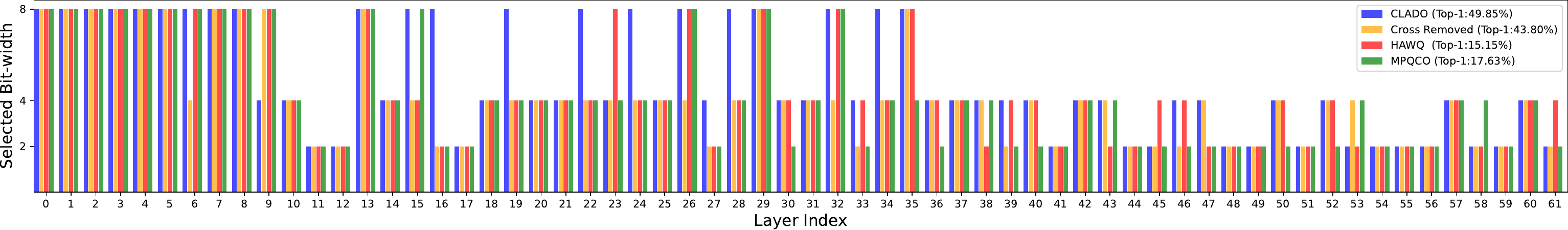}
                \caption{0.205MB}
        \end{subfigure}
        \begin{subfigure}{\textwidth}
                \centering
        	\includegraphics[width=\textwidth]{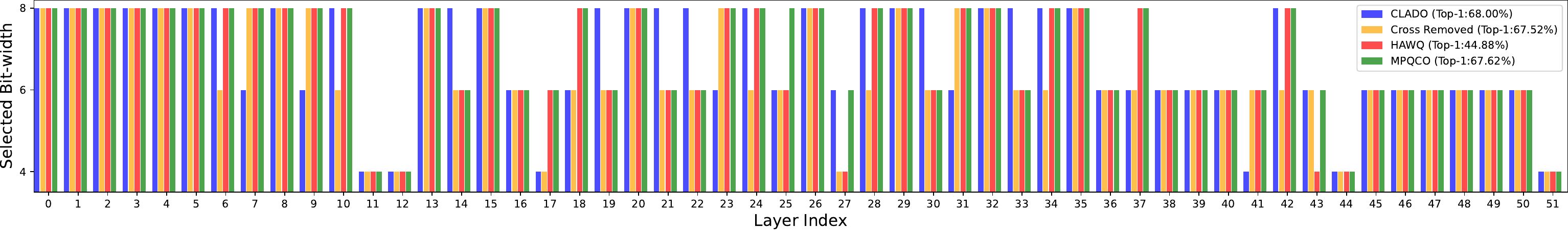}
                \caption{0.248MB}
        \end{subfigure}
     \caption{MobileNetV3}
     \label{fig:mbndec}
\end{figure}

\begin{figure}[!htb]
    \centering
        \begin{subfigure}{\textwidth}
                \centering
        	\includegraphics[width=\textwidth]{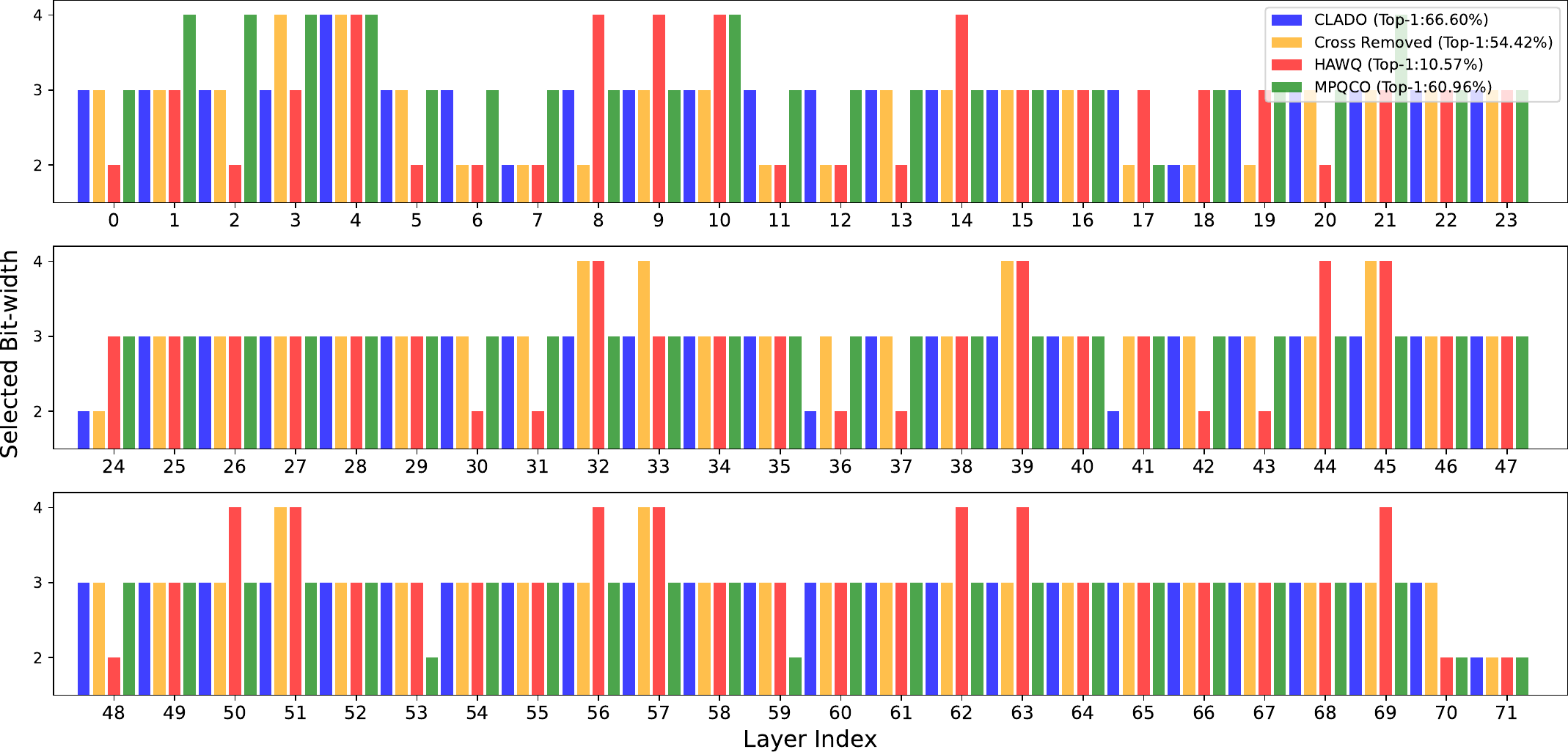}
                \caption{29.84MB}
        \end{subfigure}
        \begin{subfigure}{\textwidth}
                \centering
        	\includegraphics[width=\textwidth]{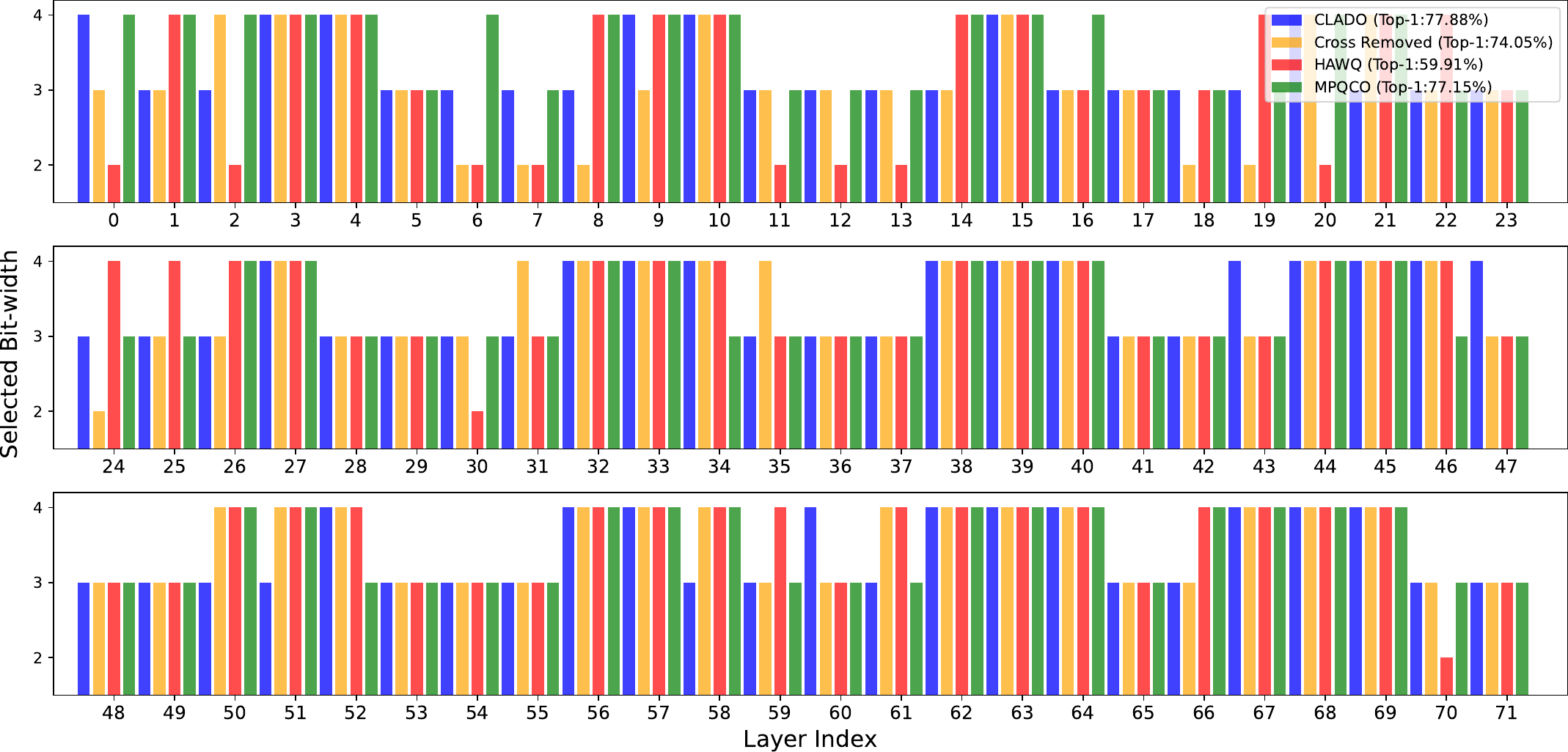}
                \caption{34.11MB}
        \end{subfigure}
     \caption{ViT}
     \label{fig:vitdec}
\end{figure}

\end{document}